\pdfoutput=1

\documentclass[11pt]{article}

\usepackage{ACL2023}

\DeclareUnicodeCharacter{0307}{u}

\usepackage{times}
\usepackage{latexsym}
\usepackage{graphicx}

\usepackage[T1]{fontenc}

\usepackage[utf8]{inputenc}

\usepackage{microtype}

\usepackage{inconsolata}
\usepackage{algorithm}
\usepackage{amsmath}
\usepackage{amsfonts}
\usepackage{amssymb}
\usepackage{amsthm}
\usepackage[noend]{algpseudocode}

\renewcommand{\algorithmiccomment}[1]{\bgroup\hfill\small//~#1\egroup}

\title{To Burst or Not to Burst: Generating and Quantifying Improbable Text}

\author{Kuleen Sasse \\
  JHUAPL\\
  \texttt{kuleen.sasse@jhuapl.edu} \\\And
  Samuel Barham \\
  JHUAPL \\
  \texttt{samuel.barham@jhuapl.edu} 
  \AND
  Efsun Sarioglu Kayi \\
  JHUAPL \\
  \texttt{efsun.kayi@jhuapl.edu} \\\And
  Edward W. Staley \\
  JHUAPL \\
  \texttt{edward.staley@jhuapl.edu} \\}

\begin{document}

\maketitle


\begin{abstract}

While large language models (LLMs) are extremely capable at text generation, their outputs are still distinguishable from human-authored text. We explore this separation across many metrics over text, many sampling techniques, many types of text data, and across two popular LLMs, LLaMA and Vicuna. Along the way, we introduce a new metric, recoverability, to highlight differences between human and machine text; and we propose a new sampling technique, burst sampling, designed to close this gap. We find that LLaMA and Vicuna have distinct distributions under many of the metrics, and that this influences our results: Recoverability separates real from fake text better than any other metric when using LLaMA. When using Vicuna, burst sampling produces text which is distributionally closer to real text compared to other sampling techniques.

\end{abstract}

\begin{figure}[!ht]
\centering
\includegraphics[width=1.05\linewidth]{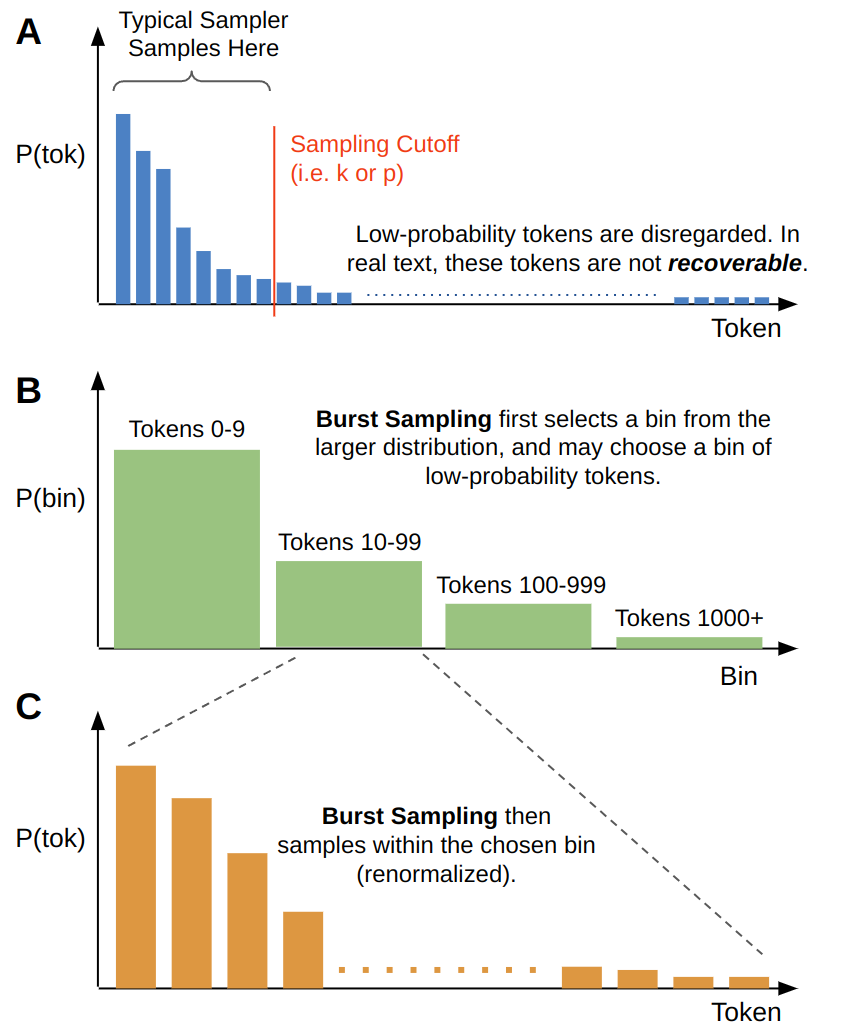}
\caption{\label{overall}Illustration of our contributions. \textbf{A} A probability-ordered token distribution with a long tail, as is commonly seen when sampling from LLMs. Sampling strategies often truncate the majority of this distribution (red line), leading to many possible tokens that cannot be sampled. When analyzing real text with an LLM, we define \emph{recoverability} as the fraction of tokens that occur before this cutoff. \textbf{B} Our \emph{burst sampling} technique first learns a distribution over bins of token ranks. To sample, a bin is first selected. \textbf{C} The probabilities of tokens within the selected bin are renormalized to form a new distribution, which is sampled. }
\end{figure}

\section{Introduction}
In recent years, large language models (LLMs) have risen as the top performing category of models for many tasks in natural language processing. Pre-trained on hundreds of millions of examples of text, these models learn probability distributions over the next token in a sequence, and these probability distributions can be used to generate text. However, while these distributions accurately describe token distributions at the corpus level, they still lead to generations which are distinguishable from human text. In this work, we explore sampling techniques used to generate text and metrics used to evaluate such text, under the lens of differences between human and machine authorship.

Our contributions in this work are threefold. \textbf{(1)} We introduce a new sampling technique called \emph{Burst Sampling}, which is designed to generate text that is statistically more similar to human-authored text than other techniques. A description of this technique can be found in Section \ref{burst}. \textbf{(2)} We introduce a new metric over a \texttt{(sampling strategy, text example)} pair called \emph{recoverability}, which captures the degree to which the given sampling strategy could have generated the text (assuming some underlying LLM providing token probabilities). This is described further in Section \ref{recoverability}. \textbf{(3)} We compute many text metrics across a variety of datasets, across real and synthetic text, across many sampling methods, and using two underlying language models. These selections are explained in Section \ref{experiments}. The results of these experiments serve as a stand-alone reference survey of common metrics and how they differ for human-authored versus generated text, and it also offers empirical justification for our recoverability metric and burst sampling technique. Our results are summarized in Section \ref{results}. Our codebase implementing these metrics and sampling techniques has been open-sourced. \footnote{https://github.com/jhuapl-fomo/burst-sampling}

\section{Related Work} \label{related}
\textbf{Text Generation.}
In this work we focus on causal language models (CLMs), which predict a distribution over next tokens given all previous tokens. This family of models is built on the transformer architecture \cite{transformer}, and was found to have performance that is proportional to model scale \cite{scaling}, leading to a growth of models from only a few hundred million parameters \cite{gpt} to tens or hundreds of billions \cite{gpt3, palm, opt}. 

These models share a similar formulation. Generation of a sequence $x$ from a probabilistic CLM with parameters $\theta$ can be defined as below:
\[
P(x_{1:T};\theta)=\prod_{t=1}^{T}P_{\theta}(x_t|x_{1:t-1})
\]

\noindent where $x_t$ is the next token, conditioned on a previous token sequence of length $t-1$.

\textbf{Sampling Strategies.}
Once the language model has produced a probability distribution over next tokens, this distribution can be sampled to generate the next piece of text. However, as this distribution encompasses tens of thousands of tokens forming the base vocabulary of the model, it has a very long tail that sums to a non-negligible probability mass. To combat against this, special sampling strategies have been devised. Deterministic decoding such as greedy selection or beam search \cite{beam} tend to produce repetitive or bland generations as they favor high probability over variety. Alternatively, sampling-based decoding methods, such as
top-k \cite{Fan2018HierarchicalNS} and top-p \cite{Holtzman2020The}, locate a nucleus of linguistically plausible tokens at the front of the distribution, and sample from these probabilistically. Other methods attempt to skew or re-weight the distribution to correct for undesirable generation artifacts, such as temperature-based sampling \cite{ACKLEY1985147}, frequency penalties \cite{ott2019fairseq}, or repetition penalties \cite{keskar2019ctrl}. Such methods can be combined as needed for finer control over the sampling process.

\textbf{Metrics for Generated Text.}
When evaluating sampling techniques and their generations, it is helpful to quantify certain aspects of the text via metrics, such as perplexity or diversity \cite{li2016diversitypromoting} measures. However, it is quite difficult to capture exactly what makes text "good" or "bad" via a direct measure, and an alternative is to measure how distinguishable generated text is from that which is truly human-authored. Among other motivations, this is a focus of synthetic text detection.

\textbf{Detecting Synthetic Text.}
With text generation becoming a widely accessible and used tool, there is widespread interest in being able to reliably detect if a piece of text was authored by a human or an LLM. Services like GPTZero \cite{gptzero} use metrics (among other factors) to help determine if text has been generated by LLMs. Meanwhile, tools like DetectGPT \cite{pmlr-v202-mitchell23a} or OpenAI's authorship classifier \cite{solaiman2019release} may be trained on specific datasets of generated text, perhaps specialized to a single target LLM. Another approach is to make synthetic text detection a design requirement of an LLM or an LLM sampling method, as seen in works on text watermarking \cite{kirchenbauer2023watermark}.

\textbf{LLM Analysis of Human Text.}
An LLM can also be used to assign probability scores to existing text rather than to generate new text. Examining LLM probability scores for real text can be used to further understand any gaps between the output of current generative models and human authors.
Previous work \citep{gehrmann-etal-2019-gltr} and \citep{Holtzman2020The} have demonstrated that tokens in human text are often not the highest probability tokens from a given language model. Especially seen in \citep{gehrmann-etal-2019-gltr}, there are regular fluctuations in LLM-provided token probabilities over the course of a human-authored piece of text, ranging from high probability to very low probability. We leverage this finding to inspire our new sampling method.


\section{Burst Sampling} \label{burst}
\subsection{Motivation} \label{EmpiricalJustification}
Intuitively, we find the design of popular sampling methods to be contradictory to the the goal of producing human-like text. In particular, there is an important distinction to be made between text that is highly probable according to an LLM, and text that is highly similar to human-authored text. LLMs which undergo pre-training are tasked with predicting which tokens are most probable given the previous context. High-probability tokens, by definition, are low-information bearing, and sampling techniques which prioritize high probability (top-p, top-k) are therefore encouraging the generation of predictable and uninformative text. In contrast to an LLM, humans author text primarily to communicate information, and therefore must include tokens that are less predictable for their audience to find value in the text itself \cite{meister, gibson}. Inspired by the probability fluctuations found in \citep{Holtzman2020The}, we introduce an algorithm, \emph{Burst Sampling}, which randomly includes tokens with high information (low probability). This is a first attempt to rectify issues with existing sampling techniques, and we hope it inspires future work that is concerned with information content in synthetic text that more closely matches human authorship.
 
\subsection{Algorithm}
As in \citep{gehrmann-etal-2019-gltr}, we divide the language model's distribution over the tokens into $n$ bins by the tokens' rank. Mimicking \citep{gehrmann-etal-2019-gltr}, each bin is between two powers of ten. For example, with 4 bins, the boundaries would be 0-10, 10-100, 100-1000, and 1000 to the end of the distribution. 

At each generation step, we sample from a categorical distribution to select a bin, and we then sample our tokens exclusively from this bin. We set all the other probabilities not in the bin to zero and normalize our distribution. This amounts to a two-tiered selection: first we select how probable our token should be, approximated by which bin we choose, and second we select a specific token from the given bin. A more in-depth explanation can be found in Algorithm \ref{algo-burst}.

The categorical distribution over bins is calculated before our generation. For any dataset or style of text we are trying to mimic, we first select a random subset of samples. Using that subset, we compute the frequency with which each bin is used: we run the model over the representative data, collect all the frequencies for each token, and assign them their corresponding rank. We then bin those values and normalize their frequencies to probabilities. 

\begin{algorithm}[t]
\caption{Burst Sampling}
\label{algo-burst}
\begin{algorithmic}[1]
\Require The sorted descending distribution over the tokens at time step $i$ represented as $P\left(x \mid x_{1:i-1}\right)$, the precalculated probabilities of each bin represented as $\boldsymbol{\theta} = (\theta_1,\theta_2,\dots,\theta_n)$ where $n$ is the number of bins, and a list of the set of indices, $\mathcal{S}$ in each of the bins represented as $\mathcal{B}$
\Ensure The modified distribution $P'\left(x \mid x_{1:i-1}\right)$
\State $b \sim Cat(\boldsymbol{\theta})$
\State $\mathcal{S} \leftarrow \mathcal{B}[b]$
\State $p' \leftarrow \sum_{x \in \mathcal{S}} P\left(x \mid x_{1:i-1}\right)$
\For {each $x$ in $P\left(x \mid x_{1:i-1}\right)$}
    \If{$x \in \mathcal{S}$}
        \State $P'\left(x \mid x_{1:i-1}\right) \leftarrow P\left(x \mid x_{1:i-1}\right)/p'$
    \Else
        \State $P'\left(x \mid x_{1:i-1}\right) \leftarrow 0$
    \EndIf
\EndFor
\end{algorithmic}
\end{algorithm}

\section{Recoverability Metric} \label{recoverability}
\subsection{Recoverability Intuition}
To further highlight and explore the differences between human-authored text and synthetic text, we introduce a new metric called \emph{recoverability}, which measures the degree to which a given sampling strategy over a given LLM could (re)produce a piece of text. The tendency of human text to periodically use low-probability tokens means that for many sampling strategies it is impossible to generate some examples of human-authored text; we say that such text is not \emph{recoverable} under the given sampling strategy. For example, a sampling strategy like top-k cannot sample any tokens which have rank>k in the LLM's output distribution. Therefore, any text using tokens with rank>k would not be recoverable under top-k.

To measure the recoverability of an entire passage of text, we measure the average recoverability of each token (what fraction of tokens are recoverable). This soft and normalized definition allows us to compare recoverability between text samples or sampling strategies.

Note that this differs from a similar metric called \textit{extractability} \cite{extractability}, which is concerned with entire sequences of tokens that have been memorized by a model and can be explicitly generated as a result of this memorization. Recoverability, by contrast, measures to what extent a sequence can be produced through the mechanism of sampling over a given distribution, and does not directly measure if such a sequence is previously known to the model.

\subsection{Mathematical Definition}
Given a nucleus function $N$ which takes a sorted descending probability distribution over tokens and partitions out the set of tokens which can be sampled (for example, top-k or top-p), and a sequence of tokens $x$ of length $T$:

\[ Recoverability(x_{1:T}) = \frac{\sum_{i=1}^{T} 1_{N(P(x_i|x_{1:i-1}))}(x_i)}{T} \]

where $1$ is the indicator function, returning 1.0 if $P(x_i)$ is in the set produced by $N$ and 0.0 otherwise. For example, if $N$ is the top-k partitioning process, then we assign 1.0 to each token within the top k tokens, and 0.0 otherwise, and then average over all these assignments.

\section{Experiments} \label{experiments}
\subsection{Overview of Experiments}
We evaluate our Burst Sampling technique and Recoverability metric as part of a larger survey over sampling strategies, metrics, and datasets, and the statistical differences that can be uncovered between human-authored and synthetic text. For each dataset of real text, we generate synthetic counterparts using selected sampling techniques, and then compute metrics over the synthetic text in comparison to the same metric computed over the real text.

\subsection{Datasets}
We consider six English-language datasets from a diverse set of domains: arXiv \citep{clement2019arxiv}, CNN/Daily Mail \citep{nallapati-etal-2016-abstractive}, Gutenberg \cite{Rae2020Compressive}, Stack Exchange \citep{pile}, Twitter \citep{rosenthal-etal-2017-semeval}, and Wikipedia \cite{wikidump}. We repeat our experiments over a variety of datasets to uncover significant differences in metrics between types of text generation.

\subsection{Language Models}
We consider LLaMA 13B \citep{touvron2023llama} as the baseline for pre-trained models. We pair that model with its fine-tuned counterpart Vicuna 13B \citep{vicuna2023}. These models are both widely used at the time of this writing, and represent the two common types of LLM that are most often used (pretrained and fine-tuned for chat). The 13B parameter models were selected to balance model size with the feasibility of such a large survey.

\subsection{Experimental Design}
Similar to \citep{Holtzman2020The}, we randomly selected 10,000 samples from each dataset to create a corresponding mini-corpus. To get samples that fit into the context windows of our models, we truncated each at 2,000 characters which is roughly 512 tokens. Since entries in the Gutenberg dataset are extremely long, we used a randomly selected paragraph from each sample in place of the full sample itself.

To generate synthetic text samples, we provide each model with the beginning of a real text sample and ask it to generate a continuation of 256 tokens. The provided real context is kept small, usually the first 10\% of the sample (the only exception being Twitter data which can have very short samples. For this dataset we used up to 5 words of the original tweet.). We computed metrics over the entire product of the generation routine (real beginning context and generated continuation), which may skew our results slightly.

Text continuation was selected as it is most appropriate for pretrained models which are fundamentally designed to continue the input text. More complex prompting blurs the line between a simple prefix and an instruction, and the latter is not appropriate for a model that is only pretrained.


\subsection{Sampling Strategies}
For each dataset and for each model, we generate synthetic text with multiple top-$k$, top-$p$, and temperature-based sampling methods. For top-$k$, we run $k$ values from $\{30,40,50\}$. For top-$p$, we generate using values from $\{0.9, 0.95, 0.99\}$, and for temperature from $\{0.5, 0.7, 0.9\}$. We additionally sample using our Burst Sampling technique as described in the previous section, in which a categorical distribution over bins is first learned for each real dataset, and is then used to select a bin to sample from at each step.

\subsection{Metrics}
For each dataset, model, and sampling technique we compute a variety of metrics over the text which can be used for classifying the text as human or synthetic, or simply to understand the text in more depth. Here we review each metric used in our analysis, in addition to the previously-described recoverability metric:



\begin{description}
    
    
    \item[Self-BLEU (diversity)] \cite{10.1145/3209978.3210080}
    For a given sentence, this metric first computes the BLEU scores (Papineni 317 et al., 2002)  between this sentence and the rest of the collection. Self-BLEU score is then calculated as the average of these scores.
    
    \item[Log Likelihood] \citep{DBLP:journals/corr/abs-1908-09203} 
    
    This approach averages the log probabilities of each token in a text. 
    
    \item[Rank and Log Rank] \citep{gehrmann-etal-2019-gltr} \citep{pmlr-v202-mitchell23a}
    Rank is calculated by finding the absolute rank for a token given its previous context. To calculate the rank score for a text sample, we average the rank for each word. To calculate the log rank, we do the same process except we sum up the tokens' log rank.
    
    \item[GLTR] \citep{gehrmann-etal-2019-gltr} introduce GLTR as a way to help distinguish whether text was generated from a language model. Its scheme of measuring the fraction of tokens that rank within a bin (0-10, 10-100, 100-1000, etc.) is a useful feature for detecting fake text as it leverages the fact that models prioritize more probable words.
    
    
    \item[Per Token K, P, and Top P Burstiness]
    
    As mentioned in Section \ref{EmpiricalJustification}, human text fluctuates frequently in probability at the token level. The current measure of burstiness included in detectors like GPTZero \cite{gptzero} does not capture that level of granularity. We instead use a per-token measure of burstiness by using the coefficient of variation for a measure. The coefficient of variation is the standard deviation of the measure divided by the mean of the measure. We propose to use the rank of the token (K value), its probability in the softmax (P), and its cumulative probability (Top P value). 
    
      
\end{description}


\subsection{Distribution Separation Measures}
For each sampling method and metric, we compute the separation between the distribution of the metric among the generated text samples to the distribution of that metric for the original text. This helps highlight, at a distributional level, differences between generated and real text under a given metric and sampling technique. For most metrics, we use the Kolmogorov–Smirnov (KS) test to measure the separation between distributions. For GLTR, which provides a metric for each bin (instead of a single value as is typical for a KS test), we train a logistic regression model to predict if a sample is real or generated from its GLTR values. We also train a logistic regression classifier using all metrics simultaneously, which can illuminate which sampling strategies lead to text that is most similar to real text when taking into account all metrics as possible decision information.

\subsection{Fluency Analysis}
Finally, to validate that our burst sampling method does not lead to text that is so random as to be "incorrect", we run a fluency analysis from UniEval \cite{zhong-etal-2022-towards} across generated text of 3,000 samples when trying to mimic the PG19, Wikipedia, and CNN Daily Mail datasets (the datasets which, subjectively, are closest to full prose). We additionally provide samples of the generated text for inspection and detailed statistics of fluency scores in the appendix (Tables \ref{burst_gens} and \ref{fluency_stats}).

\section{Results} \label{results}
\subsection{Overview of Results}




Our primary results consist of common metrics calculated over many sampling methods and across six different datasets, each with many thousands of samples. These results are extensive; please see the appendix for mean results over all our experiments. For our analysis, we also considered metrics at the distributional level, and additionally provide distance measures between real and synthetic text for each (model, dataset, sampling method, metric) combination in the appendix.

We find that there are clear differences in metrics between LLaMA and Vicuna, reflecting the tuning of the latter model. Vicuna has a notably lower perplexity than LLaMA, a higher value for k-burstiness and a lower value for p-burstiness. We further discuss the two models in \ref{model_diff}. Average metric values across our entire set of results can be found in the central columns of Table \ref{average_metrics_table}.

Our recoverability metrics are very successful in separating real and generated text using LLaMA, which we discuss in \ref{recov_results}. Additionally, our burst sampling method has a distinct effect on the distributions of samples from Vicuna, which we discuss in \ref{burst_results} below. We further validate our burst sampling method with a fluency analysis, discussed in \ref{fluency_results}.

\begin{table}
\small
\centering
\begin{tabular}{| l | c c | c c |}
\hline
 & \multicolumn{2}{|c|}{\textbf{Avg. Metric}} & \multicolumn{2}{|c|}{\textbf{Avg. Separation}}\\
\hline
\textbf{Metric} & \textbf{LLaMA} & \textbf{Vicuna} & \textbf{LLaMA} & \textbf{Vicuna}\\
\hline
k burst. & 7.869 & 9.673 & 0.341 & 0.525\\
p burst. & 0.863 & 0.577 & 0.372 & 0.813\\
top-p burst. & 0.363 & 0.241 & 0.392 & 0.206\\
log-likelihood & -1.887 & -1.211 & 0.467 & \textbf{0.855}\\
log rank & 0.851 & 0.503 & 0.458 & 0.820\\
rank & 39.978 & 39.799 & 0.317 & 0.349\\
perplexity & 9.475 & 4.736 & 0.467 & \textbf{0.855}\\
diversity & 0.795 & 0.781 & 0.433 & 0.417\\
recov top k=40 & 0.946 & 0.966 & \textbf{0.524} & 0.765\\
recov top k=50 & 0.953 & 0.970 & 0.522 & 0.743\\
\hline
\end{tabular}
\caption{\label{average_metrics_table}Average metrics for generated text over all datasets and sampling methods, and the average separation between distributions of metric values for real and generated text. In the separation columns, the \emph{highest} value is highlighted in each column, indicating that the given metric is the best, on average, at providing a distinction between real and synthetic text.}
\end{table}


\subsection{Recoverability Metric Results}\label{recov_results}
\begin{figure}[h]
\centering
\includegraphics[width=0.5\textwidth]{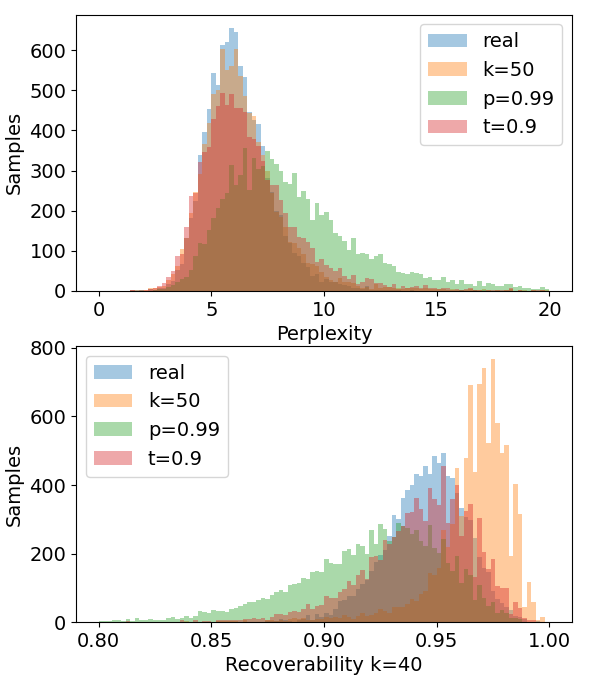}
\caption{\label{recov_hist}Distributions of perplexity and recoverability for the CNN Daily Mail dataset with several sampling methods using LLaMA. Under recoverability, the metric distributions are more separable than under other metrics.}
\end{figure}

Our recoverability metric is seen to be more successful at separating out generated text from real text when using LLaMA, but not when using Vicuna. That is, the distribution of recoverability across generated samples is more distinct from the recoverability of real text, and this separation is more pronounced than in other metrics. This can be seen in the right columns of Table \ref{average_metrics_table}, which provide the average separability of metric distributions from those of real text, averaged for each metric. This can also be visualized as distributions, as seen in Figure \ref{recov_hist} which shows example metrics' distribution under LLaMA for real text, top-k=50, top-p=0.99, temperature=0.9, and burst sampling.

It is especially interesting that our recoverability metric works well across sampling methods, because it was designed to work for a specific method at a time. This indicates that recoverability using, say, k=40, is useful for detecting synthetic text even if that text was generated with something like top-p sampling.

\subsection{Burst Sampling Results}\label{burst_results}
Our burst sampling technique tends to produce text with metric distributions closer to real text than other sampling techniques, but only when using Vicuna as the underlying model. We can see this notionally in Table \ref{average_sampler_sep_table}, which provides the average separability (KS test result) for each sampling technique, averaged over all datasets and metrics. Burst sampling is the lowest (most similar to real text) under Vicuna, but not an extreme value under LLaMA.

\begin{figure}[h]
\centering
\includegraphics[width=0.5\textwidth]{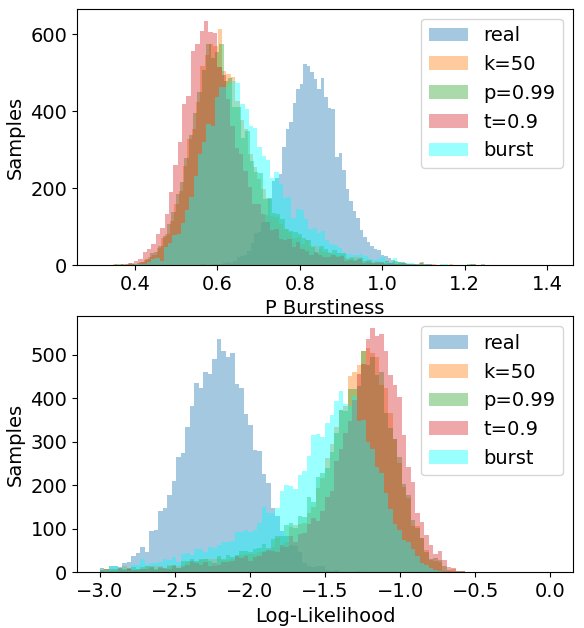}
\caption{\label{burst_hist}Distributions of P-burstiness and log-likelihood for the CNN Daily Mail dataset with several sampling methods using Vicuna. Burst sampling (turquoise) is typically closer to the distribution of real text (blue) than other metrics.}
\end{figure}

This can be further visualized as a distribution, as seen in Figure \ref{burst_hist} which shows example metrics distributions under Vicuna for real text, top-k=50, top-p=0.99, temperature=0.9, and burst sampling. The burst sampling is clearly shifted closer to the real text distribution, which we see repeatedly in our analysis.

This trend was consistent when training logistic regression classifiers on GLTR bins to consolidate them into a single separability measure. Burst sampling lead to the lowest F1 scores when used with Vicuna, indicating that it produced text which was harder to distinguish from real text when compared to other sampling methods. When using burst sampling with LLaMA, this aspect varied among the datasets. These results are given in the appendix, in Table \ref{gltr_f1}.

This trend was also consistent when running a similar logistic regression analysis using all metrics as input features (Table \ref{logreg_f1}). Burst sampling was slightly harder to detect using Vicuna, but not using LLaMA, where top-p=0.99 was clearly the hardest to distinguish. Overall, we found that these general logistic regression classifiers across all metrics performed extremely well, with F1 scores on LLaMA averaging 0.921 and on Vicuna averaging 0.986. This leads us to believe that this has merit as a general synthetic text detection mechanism.

\begin{table}[ht]
\small
\centering
\begin{tabular}{| c | c c |}
\hline
\textbf{Sampling} & \textbf{LLaMA} & \textbf{Vicuna}\\
\hline
k=30 & 0.477 & 0.642\\
k=40 & 0.447 & 0.636\\
k=50 & 0.420 & 0.634\\
p=0.9 & 0.347 & 0.630\\
p=0.95 & \textbf{0.259} & 0.640\\
p=0.99 & \textbf{0.259} & 0.599\\
t=0.5 & 0.699 & 0.723\\
t=0.7 & 0.587 & 0.696\\
t=0.9 & 0.294 & 0.638\\
burst & 0.504 & \textbf{0.512}\\
\hline
\end{tabular}
\caption{\label{average_sampler_sep_table}For each sampling method, average separation between distributions of metric values for real and generated text, over all datasets and metrics. The \emph{lowest} value is highlighted in each column, indicating that the given sampling strategy, on average, produced text that is closest to real text in terms of metrics distributions. Notation note: k and p refer to top-k and top-p sampling. t refers to temperature-based sampling.}
\end{table}

\begin{table}
\small
\centering
\begin{tabular}{| c | c c |}
\hline
\textbf{Metric} & \textbf{LLaMa} & \textbf{Vicuna}\\
\hline
real & 0.782 & 0.783\\
k=30 & 0.812 & 0.838\\
k=40 & 0.801 & 0.834\\
k=50 & 0.805 & 0.834\\
p=0.9 & 0.790 & 0.839\\
p=0.95 & 0.803 & 0.843\\
p=0.99 & 0.757 & 0.830\\
t=0.5 & 0.835 & 0.856\\
t=0.7 & 0.833 & 0.851\\
t=0.9 & 0.789 & 0.829\\
burst & 0.604 & 0.798\\
\hline
\end{tabular}
\caption{\label{fluency}For each sampling strategy, the computed average fluency score (0 to 1, 1 is most fluent) when using LLaMA and Vicuna. We limited this experiment to CNN Daily Mail, PG-19, and Wikipedia datasets.}
\end{table}

\subsection{Fluency Analysis Results}\label{fluency_results}
Our fluency analysis on samples (real and generated) from CNN Daily Mail, PG-19 and Wikipedia indicate that most sampling techniques have similar average fluency scores, and that this is similar to scores for real text. The only exception seems to be burst sampling using LLaMA, which is noticeably less fluent than other cases. This may partially explain why our burst sampling method was less effective with LLaMA- the generated text is less fluent than is typical. It is possible that for this case, our sampling strategy introduces too much random token selection, to the detriment of the generated text. It is surprising that this is not the same for both models. For fluency values across all sampling methods, please see Table \ref{fluency}.

\subsection{Model Differences}\label{model_diff}
\begin{figure}[h]
\centering
\includegraphics[width=0.5\textwidth]{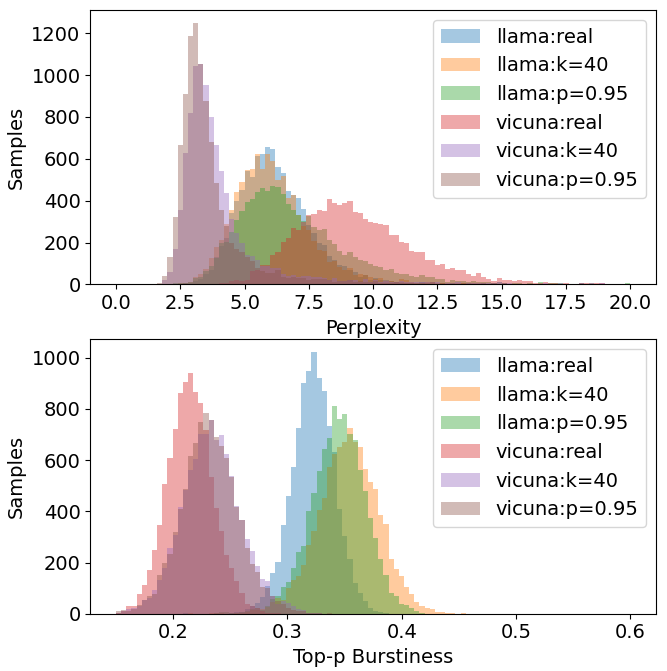}
\caption{\label{llama_vicuna_hist}Distributions of perplexity and Top-p burstiness for LLaMA and Vicuna on the CNN Daily Mail dataset. Vicuna has clear distributional differences to LLaMA on certain metrics.}
\end{figure}

Throughout our analysis, we found distinct differences between the LLaMA and Vicuna models, as discussed previously with respect to burst sampling and recoverability. Vicuna typically had higher K burstiness but lower P burstines and Top-p burstiness, as well as lower perplexity (along with associated differences in the related metrics of ranking and likelihood). However, perplexity was typically higher on Vicuna when evaluating real text. This indicates that while Vicuna is less able to generate realistic text (or at least, finds real text more perplexing than LLaMA), this does not extend to text that is generated by Vicuna itself. We attribute all of the differences to Vicuna having been specialized to chat behavior, a more narrow use-case than LLaMA, which is multi-purpose. We provide a visualization of some metric distribution distances between the two models in Figure \ref{llama_vicuna_hist}. We also provide the degree to which a given metric is different between the two models in Table \ref{model_sep}. Two of our burstiness metrics (P and Top-p) show substantial differences between the two models, and could potentially highlight if a model has been fine-tuned.

\begin{table}
\small
\centering
\begin{tabular}{| l | c |}
\hline
\textbf{Metric} & \textbf{Avg KS Between Models}\\
\hline
k burst & 0.276\\
p burst & 0.772\\
top-p burst & 0.886\\
log-likelihood & 0.685\\
log rank & 0.669\\
rank & 0.200\\
perplexity & 0.685\\
diversity & 0.209\\
recov top k=40 & 0.378\\
recov top k=50 & 0.354\\
\hline
\end{tabular}
\caption{\label{model_sep}For each metric, the average separability (KS test) over all sampling methods between the distribution when using LLaMA and that when using Vicuna.}
\end{table}

\section{Discussion} \label{discussion}


Our study over many metrics and sampling strategies, including burst sampling and recoverability, uncovered some interesting data points. A standout trend was the difference between LLaMA and Vicuna. Our hypothesis is that Vicuna sometimes produces probability distributions that are more front-weighted than LLaMA (the first portion of the sorted distribution carries more probability mass than in LLaMA). This is supported by our perplexity results- Vicuna has very low perplexity for its own generations, but high perplexity for real text. This is to be expected for a fine-tuned model that is no longer intended for general purpose NLP tasks, and it is nice to find this reflected quantitatively.

We found that burst sampling is especially helpful for Vicuna, and we believe it acts as a correcting mechanism to Vicuna's overconfidence. For LLaMA, which has a less-skewed distribution, our burst sampling is less effective since it is trained to match the distributions of real text (the same objective as LLaMA). Thus, for a purely pre-trained network, it would be a good idea to increase (or somehow calibrate) the stochasticity of burst sampling.

Recoverability is also impacted by model differences. Since we test recoverability with k=40 and k=50 (a fixed rank cutoff), we could expect that this encompasses much more probability mass in Vicuna than in LLaMA. Therefore, more of Vicuna’s generations are recoverable, and less separable from real text. However, when using LLaMA, more tokens fall beyond this threshold and it is easier to use recoverability as a separating metric for real and generated text. We were surprised that recoverability worked well even for sampling methods other than top-k, and believe it has merit as a general metric.

Finally, although we initially introduced logistic regression over all metrics to test the effectiveness of burst sampling, we actually found it to be a very reliable synthetic text classifier, especially on Vicuna. We would recommend that future synthetic text detection platforms consider this method as part of their system. Our analysis also included our burstiness metrics (K, P, and Top-p burstiness), which could certainly be further optimized for text detection by considering the entire pattern of rank or probability over a text sequence, and using time-series classification to detect fluctuations that are more or less natural. This would perhaps motivate stronger implementations of our burst sampling, to consider the time dynamics of when improbable tokens are introduced.

\section{Conclusions and Future Work} \label{conclusion}


We contributed a study of common metrics over many datasets and sampling techniques, using a pretrained model (LLaMA) and a fine-tuned model (Vicuna). Within this study we tested novel ideas of \emph{recoverability} and \emph{burst sampling}, which illuminated many interesting future directions for studying the differences between human-authored and machine generated text. We found recoverability and a logistic regression over all metrics to be promising for detecting synthetic text. Future work could further test the applicability of these results to more models and sampling methods, in particular exploring how recoverability could be used for fine-tuned models. We also found burst sampling to somewhat correct for distributional differences in Vicuna, but certainly not completely. Future work should look to calibrate and amplify burst sampling, as well as look to characterize the probability distributions between pre-trained and fine-tuned models in more detail.

\section{Ethics Statement}
We identify two possible ethical issues with our work.

First, this paper discusses the use of a sampling method that could be used to reduce the effectiveness of fake text detectors for the purposes of cheating or plagiarism. While there is a drop in detection ability for some models, the effect is fairly small and still has the ability to be detected a majority of times.

In addition, this sampling method looks at different parts of the distribution which could have the possibility of generating toxic or incorrect information. No examples of extremely toxic generations have been reported in the samples we have seen. As for misinformation, language models already generate misinformation regularly, and this method does not balloon its frequency as it only changes a few tokens per generation. 

\section{Limitations}
We identify the following limitations in our work. 

First, our analysis of different model based metrics only included two of the latest large language models. In addition, these models were in only one of the many sizes released. However, due to the model agnostic nature, we believe similar patterns to hold at different scales.  

Second, our analysis of detector performance did not use all state of the art methods like GPTZero, DetectGPT, etc. and focused on using thresholding methods. While not state of the art, it gives a picture into the possible performance of other methods as they rely on similar modelling techniques.

Third, there was considerable leakage of our datasets in the training data of LLaMA and thus Vicuna. LLaMA used arXiv, StackExchange, Books (a superset of PG19), and Wikipedia in its training set. Despite this leakage of 4 of our datasets, prompting the model with exact words from these datasets did not lead to the exact reproduction. This also can be further seen in the imperfect recoverability of the text.  

\newpage
\clearpage

\bibliography{anthology,custom,sampling_citation}
\bibliographystyle{acl_natbib}

\newpage
\clearpage
\section*{Appendix}
\section*{A.1 Burst Sampling Example Generations}
Please see Table \ref{burst_gens} for a few examples of passages generated with burst sampling.

\begin{table*}
\small
\begin{center}
\begin{tabular}{|p{8cm}  | p{8cm} |}
\hline
\textbf{LLaMA} & \textbf{Vicuna}\\
\hline
\textbf{CNN Daily Mail} & \textbf{CNN Daily Mail}\\
\hline
 War correspondent Michael Hastings may have been using marijuana to treat post-traumatic stress disorder before his death, but drug use doesn't account for the crash that killed him revered Los Angeles Times reporter. It was a tragic accident.
The car wreck that killed the war writer was caused by a blown tarp on the freeway below his car, hit him at a dangerous speed and was not caused by his driving at all, California Highway Patrol Officer Gisela Arroy said April 21". "He was driving fine. T
&
 War correspondent Michael Hastings may have been using marijuana to treat post-traumatic stress disorder before his death, but drug use doesn't appear to have been a factor (http://tinyurl.com/kvf39je) in the crash that killed him, according to the toxicology report released Tuesday.
Hastings, who wrote for Rolling Stone, BuzzFeed and other publications, was killed in a fiery car crash in Los Angeles on June 18. The coroner's office ruled the death an accident, and said Hastings appeared to hav
\\
\hline
 (CNN) -- On Tuesday, 47 Republican senators voted against a measure that would have allowed the Paycheck Fairness Act to go to a vote in the Senate. Why not let the bill go to a full vote? The Republican Party didn't want to confront the "gender wage gap" head long.
The 1991 Lilly Ledbetter Fair Pay Act was a huge step forward in the fight for worker parity. The new bill takes that act a step further, and is supported by a majority of the US Senate. This bill has also been backed widely by wome
&
 (CNN) -- On Tuesday, 47 Republican senators voted against a measure that would have allowed the Paycheck Fairness Act to go to a vote in the Senate. Why not let the bill go to a full vote? The Republican Party didn't want to confront the issue of gender pay inequality. They were more interested in protecting their donors, who they know would not be happy about the Paycheck Fairness Act.

The Paycheck Fairness Act would have helped to close the wage gap between men and women. It would have provi
\\
\hline
\textbf{PG-19} & \textbf{PG-19}\\
\hline
 They went slowly down the hill, their heads bent against the driving rain. They were not thinking about the recurring arguments re the need discovered of new houses but the new houses which they had built, their own homes, their own possession.
Their thoughts were of home, busy after a hard days work on Earl Hood’s farm, a day doubly length but not a moment of which was spent in waste. It had been a day of satisfaction but Arlene was thinking at the moment there was more satisfaction in the tho
&
 They went slowly down the hill, their heads bent under the heavy burden, their steps lagging wearily behind the eager prisoners, who were impatient at the slow progress.

The novelty of so many high-born ladies and knights as prisoners had not quite worn off, and the curiosity of the people was as keen as ever. Sir EDWARD, whose armor was the finest and most becoming of the number, was surrounded by a fair group of admirers, especially young ladies who could fan themselves with his plumes, and 
\\
\hline
 Altogether, that evening stood out in Ruth Erskine Tyler Sill's life as one exemplifying the truth of her favorite saying, “Life's a gift. Beware of expectoritions”-- a saying that her brother, a young practising physincin, had originated from the Latin, and then, when it had become the text of an address in that language, had rumbled up and down to the astonishment of the facolty.

Ruth's father was a lawyer of importance, and, in addition to that, a member.

“Of the legislature?” Ruth'she bro
&
 Altogether, that evening stood out in Ruth Erskine's mind as a period of deep loneliness.

As she sat by the solitary candle, thinking of her absent lover, and of the friends who had deserted her, her thoughts grew more and more despondent. She began to question whether she was wise (she wondered) to have refused Mr. Hardcondé, and to have placed so much reliance on the uncertain future. For the time, life appeared very dark to her.

But the darkness lifted. The fretful mood passed away, and a 
\\
\hline
\textbf{Wikipedia} & \textbf{Wikipedia}\\
\hline
 The Woolwich pub bombing was an attack by the Provisional Irish Republican Army in Woolwich, a district of London in the United Kingdom, as part of The Troubles.

Echoing similar bombings in Dublin, the Woolwich Arsène attack took place fifty yards away from the Royal Artillery Barracks in the Eltham Hill New Town area on the morning of Wednesday June 9, 1973 when three Irish Republicans planted a time bomb at the rear of the barracks, which had been evacuated earlier in the day.

At 10:$\pi$:0
&
 The Woolwich pub bombing was an attack by the Provisional Irish Republican Army in Woolwich, a district of London in the United Kingdom, as part of The Troubles.

Echoing similar attacks in the Basque Country and Italy, the IRA carried out the Woolwich bombing in an attempt to disrupt the Anglo-Republican Treaty and sow chaos in London.

On February 20, 1974, two car bombs were detonated outside the St. George's Inn pub in Woolwich, killing four people and inflicting dozens of injuries. The att
\\
\hline
 Zornia glabra is a species of legume (Fabaceae), native to Africa, Australia, Madagascar and the Arab ... Many other species were once included in the Genus and are now included in ...
Zornia latifolia is a species of legume (Fabaceae), widely used in horticulture in the N... It is sometimes included in the Genistorus subgenus of Astragalus.
Zornia latifolia is a species of leg..
Physiological responses to water stress in zornia latitifo..
Agriculture is a vital part of human life. In India agr
&
 Zornia glabra is a species of flowering plant in the legume family known by several common names, including smooth Apache plume, smooth sicklepod, sicklepod, and devil's shoestring. It is native to several countries in the Americas, including the United States, Canada, Mexico, and several Caribbean islands. It is an annual or perennial herb that can grow to a height of about 3 meters (10 feet). The plant produces clusters of small, yellow flowers and long, thin pods that contain seeds. It is of
\\
\hline
\end{tabular}
\caption{\label{burst_gens}Randomly selected generations from burst sampling, continuing text from CNN Daily Mail, PG-19, and Wikipedia. Samples are clipped for display purposes.}
\end{center}
\end{table*}

\section*{A.2 All Metrics}

We provide tables of all our metric results and separability tests, broken down by dataset. Please see the tables in the subsequent pages.

\begin{table*}
\small
\centering
\begin{tabular}{|c|c|cccc|cccc|}
\hline
&&\multicolumn{4}{|c|}{\textbf{LlaMa}}&	\multicolumn{4}{|c|}{\textbf{Vicuna}}	\\
\hline
\textbf{Metric}&\textbf{Count}&\textbf{Min}&\textbf{Max}&\textbf{Mean}&\textbf{Variance}&\textbf{Min}&\textbf{Max}&\textbf{Mean}&\textbf{Variance} \\
\hline
real&3000&0.006&0.971&0.782&0.034&0.018&0.967&0.783&0.033 \\
k=30&3000&0.020&0.970&0.812&0.027&0.004&0.971&0.838&0.027 \\
k=40&3000&0.003&0.969&0.801&0.030&0.003&0.971&0.834&0.028 \\
k=50&3000&0.013&0.967&0.805&0.028&0.012&0.976&0.834&0.028 \\
p=0.9&3000&0.003&0.969&0.790&0.032&0.025&0.970&0.839&0.026 \\
p=0.95&3000&0.014&0.969&0.803&0.028&0.004&0.972&0.843&0.025 \\
p=0.99&3000&0.006&0.969&0.757&0.036&0.004&0.974&0.830&0.030 \\
t=0.5&3000&0.003&0.976&0.835&0.024&0.011&0.971&0.856&0.022 \\
t=0.7&3000&0.011&0.967&0.833&0.023&0.003&0.974&0.851&0.023 \\
t=0.9&3000&0.026&0.969&0.789&0.031&0.008&0.970&0.829&0.030 \\
burst&3000&0.003&0.968&0.604&0.042&0.009&0.969&0.798&0.033 \\
\hline
\end{tabular}
\caption{\label{fluency_stats}Statistics of fluency score (0 to 1, 1 is most fluent) when using LLaMA and Vicuna.}
\end{table*}

\begin{table*}
\small
\begin{center}
\begin{tabular}{| l | c c c c c c c c c c c |}
\hline
\multicolumn{12}{|c|}{\textbf{Average arXiv Metrics}}\\
\hline
\textbf{LLaMA 13B}& & & & & & & & & & & \\
\hline
Sampling & real & k=30 & k=40 & k=50 & p=0.9 & p=0.95 & p=0.99 & t=0.5 & t=0.7 & t=0.9 & burst\\
\hline
k burst & 5.99 & 8.63 & 8.36 & 8.13 & 7.45 & 6.65 & 5.79 & 9.74 & 8.65 & 6.87 & 7.4 \\
p burst & 0.92 & 0.84 & 0.86 & 0.87 & 0.84 & 0.89 & 0.95 & 0.6 & 0.71 & 0.86 & 1.07 \\
top-p burst & 0.33 & 0.36 & 0.35 & 0.35 & 0.34 & 0.34 & 0.35 & 0.39 & 0.37 & 0.36 & 0.41 \\
log-likelihood & -2.24 & -1.72 & -1.77 & -1.82 & -1.78 & -1.98 & -2.25 & -1.09 & -1.39 & -1.92 & -2.75 \\
log rank & 1.06 & 0.73 & 0.77 & 0.8 & 0.81 & 0.94 & 1.1 & 0.37 & 0.55 & 0.88 & 1.21 \\
rank & 41.07 & 16.97 & 17.33 & 17.8 & 20.54 & 24.36 & 35.79 & 15.78 & 17.91 & 28.85 & 72.93 \\
perplexity & 10.36 & 5.75 & 6.1 & 6.39 & 6.17 & 7.78 & 10.32 & 3.08 & 4.15 & 7.23 & 16.15 \\
diversity: & 0.84 & 0.79 & 0.8 & 0.8 & 0.8 & 0.81 & 0.83 & 0.64 & 0.73 & 0.8 & 0.83 \\
recov top k=40 & 0.91 & 0.98 & 0.98 & 0.97 & 0.95 & 0.93 & 0.91 & 0.98 & 0.97 & 0.94 & 0.91 \\
recov top k=50 & 0.92 & 0.98 & 0.98 & 0.98 & 0.96 & 0.94 & 0.92 & 0.98 & 0.97 & 0.94 & 0.92 \\
\hline
\textbf{Vicuna 13B}& & & & & & & & & & & \\
\hline
k burst & 5.77 & 10.21 & 10.13 & 10.08 & 9.42 & 10.04 & 9.42 & 10.9 & 10.64 & 9.85 & 8.44 \\
p burst & 0.93 & 0.6 & 0.6 & 0.61 & 0.62 & 0.58 & 0.62 & 0.45 & 0.5 & 0.58 & 0.69 \\
top-p burst & 0.23 & 0.24 & 0.24 & 0.24 & 0.24 & 0.24 & 0.24 & 0.25 & 0.24 & 0.24 & 0.25 \\
log-likelihood & -2.61 & -1.22 & -1.24 & -1.26 & -1.32 & -1.16 & -1.32 & -0.83 & -0.94 & -1.19 & -1.66 \\
log rank & 1.19 & 0.5 & 0.51 & 0.52 & 0.57 & 0.49 & 0.57 & 0.3 & 0.36 & 0.49 & 0.71 \\
rank & 59.24 & 32.62 & 33.03 & 33.13 & 36.45 & 32.1 & 36.45 & 28.77 & 30.08 & 34.43 & 52.89 \\
perplexity & 15.19 & 3.73 & 3.83 & 3.9 & 4.23 & 3.52 & 4.23 & 2.43 & 2.74 & 3.63 & 6.08 \\
diversity: & 0.84 & 0.77 & 0.78 & 0.78 & 0.78 & 0.77 & 0.78 & 0.72 & 0.74 & 0.77 & 0.81 \\
recov top k=40 & 0.89 & 0.97 & 0.97 & 0.97 & 0.96 & 0.97 & 0.96 & 0.98 & 0.98 & 0.97 & 0.95 \\
recov top k=50 & 0.9 & 0.98 & 0.98 & 0.97 & 0.97 & 0.97 & 0.97 & 0.98 & 0.98 & 0.97 & 0.95 \\
\hline
\end{tabular}
\caption{\label{arxiv_avg}Average metrics for the arXiv dataset, for each sampling method and each model. Notation note: k and p refer to top-k and top-p sampling. t refers to temperature-based sampling.}
\end{center}
\end{table*}

\begin{table*}
\small
\begin{center}
\begin{tabular}{| l | c c c c c c c c c c c |}
\hline
\multicolumn{12}{|c|}{\textbf{Average CNN Daily Mail Metrics}}\\
\hline
\textbf{LLaMA 13B}& & & & & & & & & & & \\
\hline
Sampling & real & k=30 & k=40 & k=50 & p=0.9 & p=0.95 & p=0.99 & t=0.5 & t=0.7 & t=0.9 & burst\\
\hline
k burst & 7.4 & 8.73 & 8.54 & 8.38 & 6.52 & 7.42 & 6.52 & 10.0 & 9.1 & 7.49 & 8.24 \\
p burst & 0.82 & 0.85 & 0.86 & 0.87 & 0.92 & 0.88 & 0.92 & 0.66 & 0.74 & 0.85 & 0.98 \\
top-p burst & 0.32 & 0.36 & 0.35 & 0.35 & 0.35 & 0.34 & 0.35 & 0.4 & 0.38 & 0.36 & 0.4 \\
log-likelihood & -1.8 & -1.75 & -1.79 & -1.82 & -2.11 & -1.9 & -2.11 & -1.2 & -1.45 & -1.86 & -2.38 \\
log rank & 0.83 & 0.77 & 0.8 & 0.82 & 1.02 & 0.9 & 1.02 & 0.43 & 0.6 & 0.86 & 1.03 \\
rank & 22.61 & 24.48 & 24.87 & 25.32 & 38.93 & 29.22 & 38.93 & 20.35 & 22.96 & 32.37 & 71.68 \\
perplexity & 6.21 & 5.95 & 6.23 & 6.44 & 8.79 & 7.02 & 8.79 & 3.42 & 4.41 & 6.72 & 11.13 \\
diversity: & 0.75 & 0.8 & 0.81 & 0.81 & 0.84 & 0.82 & 0.84 & 0.7 & 0.76 & 0.81 & 0.83 \\
recov top k=40 & 0.95 & 0.97 & 0.97 & 0.97 & 0.92 & 0.94 & 0.92 & 0.98 & 0.97 & 0.94 & 0.93 \\
recov top k=50 & 0.95 & 0.98 & 0.98 & 0.97 & 0.93 & 0.95 & 0.93 & 0.98 & 0.97 & 0.95 & 0.94 \\
\hline
\textbf{Vicuna 13B}& & & & & & & & & & & \\
\hline
k burst & 7.7 & 9.92 & 9.85 & 9.84 & 10.01 & 10.01 & 9.67 & 10.74 & 10.45 & 9.89 & 9.07 \\
p burst & 0.83 & 0.62 & 0.63 & 0.63 & 0.6 & 0.6 & 0.63 & 0.5 & 0.54 & 0.6 & 0.67 \\
top-p burst & 0.21 & 0.23 & 0.23 & 0.23 & 0.23 & 0.23 & 0.23 & 0.25 & 0.24 & 0.24 & 0.24 \\
log-likelihood & -2.2 & -1.35 & -1.37 & -1.37 & -1.27 & -1.27 & -1.38 & -0.99 & -1.1 & -1.3 & -1.57 \\
log rank & 0.93 & 0.56 & 0.57 & 0.57 & 0.53 & 0.53 & 0.58 & 0.37 & 0.44 & 0.54 & 0.66 \\
rank & 33.84 & 43.0 & 43.58 & 43.33 & 41.11 & 41.11 & 43.77 & 35.07 & 37.55 & 42.23 & 51.43 \\
perplexity & 9.34 & 4.28 & 4.39 & 4.4 & 3.93 & 3.93 & 4.44 & 2.82 & 3.21 & 4.05 & 5.39 \\
diversity: & 0.75 & 0.8 & 0.8 & 0.8 & 0.79 & 0.79 & 0.8 & 0.76 & 0.77 & 0.79 & 0.82 \\
recov top k=40 & 0.93 & 0.97 & 0.97 & 0.97 & 0.97 & 0.97 & 0.96 & 0.97 & 0.97 & 0.96 & 0.95 \\
recov top k=50 & 0.94 & 0.97 & 0.97 & 0.97 & 0.97 & 0.97 & 0.97 & 0.98 & 0.97 & 0.97 & 0.96 \\
\hline
\end{tabular}
\caption{\label{cnn_dailymail_avg}Average metrics for the CNN Daily Mail dataset, for each sampling method and each model.}
\end{center}
\end{table*}

\begin{table*}
\small
\begin{center}
\begin{tabular}{| l | c c c c c c c c c c c |}
\hline
\multicolumn{12}{|c|}{\textbf{Average PG19 Metrics}}\\
\hline
\textbf{LLaMA 13B}& & & & & & & & & & & \\
\hline
Sampling & real & k=30 & k=40 & k=50 & p=0.9 & p=0.95 & p=0.99 & t=0.5 & t=0.7 & t=0.9 & burst\\
\hline
k burst & 4.86 & 7.97 & 7.62 & 7.38 & 5.86 & 5.86 & 5.16 & 10.17 & 8.45 & 6.34 & 6.71 \\
p burst & 1.04 & 0.95 & 0.97 & 0.98 & 1.01 & 1.01 & 1.08 & 0.59 & 0.76 & 0.96 & 1.23 \\
top-p burst & 0.36 & 0.39 & 0.39 & 0.38 & 0.37 & 0.37 & 0.37 & 0.39 & 0.4 & 0.39 & 0.46 \\
log-likelihood & -2.64 & -1.94 & -2.02 & -2.06 & -2.27 & -2.27 & -2.6 & -1.05 & -1.5 & -2.18 & -3.13 \\
log rank & 1.28 & 0.84 & 0.89 & 0.93 & 1.11 & 1.11 & 1.31 & 0.34 & 0.59 & 1.03 & 1.42 \\
rank & 55.02 & 18.08 & 19.13 & 19.54 & 30.09 & 30.09 & 49.26 & 15.27 & 18.6 & 36.17 & 99.79 \\
perplexity & 16.38 & 7.32 & 8.07 & 8.37 & 10.69 & 10.69 & 15.1 & 3.05 & 4.8 & 9.65 & 23.74 \\
diversity: & 0.88 & 0.79 & 0.8 & 0.8 & 0.82 & 0.82 & 0.84 & 0.59 & 0.72 & 0.81 & 0.83 \\
recov top k=40 & 0.89 & 0.98 & 0.98 & 0.97 & 0.92 & 0.92 & 0.88 & 0.98 & 0.97 & 0.92 & 0.88 \\
recov top k=50 & 0.9 & 0.98 & 0.98 & 0.98 & 0.93 & 0.93 & 0.9 & 0.98 & 0.97 & 0.93 & 0.89 \\
\hline
\textbf{Vicuna 13B}& & & & & & & & & & & \\
\hline
k burst & 4.89 & 9.8 & 9.67 & 9.57 & 9.83 & 9.41 & 8.73 & 10.84 & 10.37 & 9.35 & 7.91 \\
p burst & 1.05 & 0.61 & 0.62 & 0.62 & 0.56 & 0.6 & 0.65 & 0.44 & 0.5 & 0.59 & 0.73 \\
top-p burst & 0.25 & 0.26 & 0.26 & 0.26 & 0.26 & 0.26 & 0.26 & 0.28 & 0.27 & 0.27 & 0.28 \\
log-likelihood & -2.98 & -1.19 & -1.22 & -1.24 & -1.06 & -1.18 & -1.35 & -0.76 & -0.9 & -1.18 & -1.71 \\
log rank & 1.41 & 0.48 & 0.49 & 0.51 & 0.43 & 0.49 & 0.58 & 0.26 & 0.33 & 0.48 & 0.74 \\
rank & 79.1 & 22.66 & 22.98 & 23.25 & 23.07 & 23.77 & 27.86 & 21.29 & 21.73 & 25.25 & 40.94 \\
perplexity & 23.63 & 3.58 & 3.8 & 3.86 & 3.28 & 3.68 & 4.45 & 4.13 & 2.73 & 3.67 & 6.49 \\
diversity: & 0.88 & 0.78 & 0.79 & 0.79 & 0.77 & 0.78 & 0.8 & 0.72 & 0.75 & 0.78 & 0.82 \\
recov top k=40 & 0.87 & 0.98 & 0.98 & 0.98 & 0.98 & 0.97 & 0.96 & 0.98 & 0.98 & 0.97 & 0.94 \\
recov top k=50 & 0.88 & 0.98 & 0.98 & 0.98 & 0.98 & 0.97 & 0.97 & 0.98 & 0.98 & 0.97 & 0.95 \\
\hline
\end{tabular}
\caption{\label{pg19_avg}Average metrics for the PG19 dataset, for each sampling method and each model.}
\end{center}
\end{table*}

\begin{table*}
\small
\begin{center}
\begin{tabular}{| l | c c c c c c c c c c c |}
\hline
\multicolumn{12}{|c|}{\textbf{Average StackExchange Metrics}}\\
\hline
\textbf{LLaMA 13B}& & & & & & & & & & & \\
\hline
Sampling & real & k=30 & k=40 & k=50 & p=0.9 & p=0.95 & p=0.99 & t=0.5 & t=0.7 & t=0.9 & burst\\
\hline
k burst & 7.93 & 8.46 & 8.0 & 7.87 & 7.72 & 7.14 & 6.32 & 8.93 & 8.37 & 7.23 & 8.18 \\
p burst & 0.79 & 0.79 & 0.82 & 0.82 & 0.77 & 0.82 & 0.88 & 0.58 & 0.66 & 0.79 & 0.97 \\
top-p burst & 0.31 & 0.34 & 0.34 & 0.34 & 0.33 & 0.33 & 0.33 & 0.36 & 0.35 & 0.34 & 0.38 \\
log-likelihood & -1.78 & -1.59 & -1.71 & -1.74 & -1.59 & -1.76 & -2.01 & -1.07 & -1.3 & -1.71 & -2.41 \\
log rank & 0.81 & 0.69 & 0.75 & 0.78 & 0.73 & 0.82 & 0.96 & 0.4 & 0.54 & 0.78 & 1.04 \\
rank & 25.92 & 18.89 & 21.92 & 22.31 & 23.0 & 25.18 & 33.4 & 19.92 & 21.5 & 28.57 & 76.04 \\
perplexity & 6.78 & 5.25 & 5.84 & 6.08 & 5.32 & 6.37 & 8.36 & 3.02 & 3.85 & 6.02 & 11.39 \\
diversity: & 0.78 & 0.81 & 0.81 & 0.81 & 0.8 & 0.81 & 0.83 & 0.7 & 0.76 & 0.81 & 0.83 \\
recov top k=40 & 0.95 & 0.98 & 0.97 & 0.97 & 0.96 & 0.95 & 0.93 & 0.97 & 0.97 & 0.95 & 0.93 \\
recov top k=50 & 0.95 & 0.98 & 0.97 & 0.97 & 0.96 & 0.95 & 0.93 & 0.98 & 0.97 & 0.95 & 0.93 \\
\hline
\textbf{Vicuna 13B}& & & & & & & & & & & \\
\hline
k burst & 7.83 & 9.01 & 8.94 & 8.97 & 9.32 & 9.15 & 8.86 & 9.61 & 9.46 & 9.03 & 8.17 \\
p burst & 0.81 & 0.58 & 0.59 & 0.58 & 0.52 & 0.55 & 0.58 & 0.45 & 0.49 & 0.55 & 0.65 \\
top-p burst & 0.21 & 0.22 & 0.22 & 0.22 & 0.22 & 0.22 & 0.22 & 0.23 & 0.22 & 0.22 & 0.23 \\
log-likelihood & -2.18 & -1.23 & -1.26 & -1.24 & -1.05 & -1.12 & -1.24 & -0.89 & -0.98 & -1.16 & -1.54 \\
log rank & 0.92 & 0.51 & 0.53 & 0.52 & 0.44 & 0.47 & 0.53 & 0.35 & 0.39 & 0.49 & 0.65 \\
rank & 39.49 & 27.7 & 28.62 & 27.94 & 25.52 & 26.16 & 28.5 & 23.96 & 24.86 & 28.14 & 38.41 \\
perplexity & 10.51 & 4.01 & 4.22 & 4.08 & 3.21 & 3.51 & 4.1 & 2.62 & 2.91 & 3.73 & 5.76 \\
diversity: & 0.78 & 0.75 & 0.76 & 0.76 & 0.74 & 0.74 & 0.75 & 0.71 & 0.73 & 0.75 & 0.77 \\
recov top k=40 & 0.93 & 0.97 & 0.97 & 0.97 & 0.97 & 0.97 & 0.96 & 0.97 & 0.97 & 0.97 & 0.95 \\
recov top k=50 & 0.94 & 0.97 & 0.97 & 0.97 & 0.97 & 0.97 & 0.97 & 0.97 & 0.97 & 0.97 & 0.96 \\
\hline
\end{tabular}
\caption{\label{stackexchange_avg}Average metrics for the StackExchange dataset, for each sampling method and each model.}
\end{center}
\end{table*}

\begin{table*}
\small
\begin{center}
\begin{tabular}{| l | c c c c c c c c c c c |}
\hline
\multicolumn{12}{|c|}{\textbf{Average Twitter Metrics}}\\
\hline
\textbf{LLaMA 13B}& & & & & & & & & & & \\
\hline
Sampling & real & k=30 & k=40 & k=50 & p=0.9 & p=0.95 & p=0.99 & t=0.5 & t=0.7 & t=0.9 & burst\\
\hline
k burst & 3.13 & 7.85 & 7.56 & 7.32 & 6.72 & 6.02 & 5.29 & 10.37 & 8.56 & 6.44 & 4.48 \\
p burst & 1.32 & 0.96 & 0.98 & 0.99 & 0.98 & 1.04 & 1.12 & 0.56 & 0.75 & 0.99 & 1.67 \\
top-p burst & 0.38 & 0.39 & 0.39 & 0.39 & 0.37 & 0.37 & 0.38 & 0.38 & 0.4 & 0.39 & 0.5 \\
log-likelihood & -3.77 & -2.03 & -2.11 & -2.17 & -2.18 & -2.43 & -2.76 & -0.97 & -1.5 & -2.3 & -4.47 \\
log rank & 1.9 & 0.9 & 0.95 & 0.99 & 1.05 & 1.21 & 1.42 & 0.32 & 0.61 & 1.11 & 2.31 \\
rank & 135.45 & 40.22 & 41.11 & 42.0 & 48.07 & 56.63 & 78.28 & 25.33 & 34.11 & 61.19 & 307.99 \\
perplexity & 64.8 & 10.57 & 11.2 & 11.92 & 13.22 & 16.52 & 22.59 & 3.15 & 5.86 & 14.24 & 94.75 \\
diversity: & 0.98 & 0.82 & 0.82 & 0.83 & 0.83 & 0.85 & 0.86 & 0.55 & 0.73 & 0.83 & 0.89 \\
recov top k=40 & 0.8 & 0.97 & 0.97 & 0.96 & 0.92 & 0.9 & 0.86 & 0.98 & 0.96 & 0.91 & 0.75 \\
recov top k=50 & 0.82 & 0.97 & 0.97 & 0.97 & 0.93 & 0.91 & 0.88 & 0.98 & 0.97 & 0.92 & 0.77 \\
\hline
\textbf{Vicuna 13B}& & & & & & & & & & & \\
\hline
k burst & 3.19 & 9.24 & 9.17 & 9.09 & 8.4 & 8.92 & 8.4 & 10.4 & 9.87 & 8.93 & 6.62 \\
p burst & 1.34 & 0.61 & 0.62 & 0.62 & 0.65 & 0.61 & 0.65 & 0.44 & 0.5 & 0.6 & 0.86 \\
top-p burst & 0.27 & 0.26 & 0.26 & 0.26 & 0.26 & 0.25 & 0.26 & 0.27 & 0.26 & 0.26 & 0.29 \\
log-likelihood & -4.14 & -1.27 & -1.29 & -1.32 & -1.44 & -1.27 & -1.44 & -0.83 & -0.98 & -1.27 & -2.35 \\
log rank & 2.1 & 0.54 & 0.55 & 0.57 & 0.65 & 0.57 & 0.65 & 0.31 & 0.4 & 0.56 & 1.06 \\
rank & 203.62 & 53.87 & 53.32 & 55.86 & 62.22 & 56.61 & 62.22 & 44.97 & 49.12 & 57.49 & 103.03 \\
perplexity & 92.06 & 7.95 & 8.11 & 9.65 & 9.26 & 8.24 & 9.26 & 5.33 & 6.69 & 9.89 & 17.87 \\
diversity: & 0.98 & 0.82 & 0.82 & 0.82 & 0.83 & 0.82 & 0.83 & 0.75 & 0.78 & 0.82 & 0.88 \\
recov top k=40 & 0.76 & 0.97 & 0.97 & 0.96 & 0.95 & 0.96 & 0.95 & 0.97 & 0.97 & 0.96 & 0.9 \\
recov top k=50 & 0.79 & 0.97 & 0.97 & 0.97 & 0.95 & 0.96 & 0.95 & 0.98 & 0.97 & 0.96 & 0.91 \\
\hline
\end{tabular}
\caption{\label{twitter_avg}Average metrics for the Twitter dataset, for each sampling method and each model.}
\end{center}
\end{table*}

\begin{table*}
\small
\begin{center}
\begin{tabular}{| l | c c c c c c c c c c c |}
\hline
\multicolumn{12}{|c|}{\textbf{Average Wikipedia Metrics}}\\
\hline
\textbf{LLaMA 13B}& & & & & & & & & & & \\
\hline
Sampling & real & k=30 & k=40 & k=50 & p=0.9 & p=0.95 & p=0.99 & t=0.5 & t=0.7 & t=0.9 & burst\\
\hline
k burst & 7.23 & 9.92 & 9.74 & 9.62 & 8.77 & 8.77 & 7.86 & 10.78 & 10.14 & 8.73 & 8.85 \\
p burst & 0.72 & 0.73 & 0.74 & 0.75 & 0.74 & 0.74 & 0.79 & 0.55 & 0.62 & 0.73 & 0.89 \\
top-p burst & 0.29 & 0.32 & 0.32 & 0.32 & 0.31 & 0.31 & 0.32 & 0.34 & 0.33 & 0.32 & 0.37 \\
log-likelihood & -1.64 & -1.5 & -1.54 & -1.56 & -1.58 & -1.58 & -1.79 & -1.01 & -1.19 & -1.56 & -2.23 \\
log rank & 0.72 & 0.63 & 0.66 & 0.67 & 0.72 & 0.72 & 0.83 & 0.37 & 0.48 & 0.69 & 0.92 \\
rank & 32.77 & 34.73 & 35.87 & 35.56 & 38.61 & 38.61 & 45.76 & 34.88 & 35.75 & 42.04 & 97.8 \\
perplexity & 5.65 & 4.81 & 5.63 & 5.18 & 5.52 & 5.52 & 6.94 & 2.96 & 3.55 & - & 9.92 \\
diversity: & 0.8 & 0.81 & 0.81 & 0.81 & 0.81 & 0.81 & 0.82 & 0.73 & 0.77 & 0.81 & 0.82 \\
recov top k=40 & 0.95 & 0.98 & 0.98 & 0.97 & 0.95 & 0.95 & 0.94 & 0.98 & 0.97 & 0.95 & 0.94 \\
recov top k=50 & 0.95 & 0.98 & 0.98 & 0.98 & 0.96 & 0.96 & 0.94 & 0.98 & 0.97 & 0.96 & 0.94 \\
\hline
\textbf{Vicuna 13B}& & & & & & & & & & & \\
\hline
k burst & 7.38 & 10.8 & 10.73 & 10.71 & 10.43 & 10.71 & 10.43 & 11.27 & 11.04 & 10.64 & 9.97 \\
p burst & 0.73 & 0.53 & 0.53 & 0.53 & 0.53 & 0.51 & 0.53 & 0.44 & 0.47 & 0.51 & 0.57 \\
top-p burst & 0.2 & 0.21 & 0.21 & 0.21 & 0.21 & 0.21 & 0.21 & 0.23 & 0.22 & 0.21 & 0.22 \\
log-likelihood & -1.97 & -1.08 & -1.09 & -1.09 & -1.11 & -1.03 & -1.11 & -0.84 & -0.92 & -1.05 & -1.26 \\
log rank & 0.81 & 0.43 & 0.44 & 0.44 & 0.46 & 0.42 & 0.46 & 0.31 & 0.36 & 0.42 & 0.51 \\
rank & 40.81 & 46.45 & 48.17 & 48.47 & 50.35 & 48.31 & 50.35 & 44.11 & 45.49 & 48.38 & 54.47 \\
perplexity & 8.0 & 3.54 & 3.56 & 3.61 & 3.84 & 3.56 & 3.84 & 2.52 & 2.95 & 3.26 & 4.23 \\
diversity: & 0.8 & 0.79 & 0.79 & 0.79 & 0.79 & 0.79 & 0.79 & 0.76 & 0.77 & 0.79 & 0.8 \\
recov top k=40 & 0.94 & 0.97 & 0.97 & 0.97 & 0.97 & 0.97 & 0.97 & 0.98 & 0.97 & 0.97 & 0.96 \\
recov top k=50 & 0.94 & 0.98 & 0.98 & 0.98 & 0.97 & 0.97 & 0.97 & 0.98 & 0.98 & 0.97 & 0.97 \\
\hline
\end{tabular}
\caption{\label{wikipedia_avg}Average metrics for the Wikipedia dataset, for each sampling method and each model.}
\end{center}
\end{table*}

\begin{table*}
\small
\begin{center}
\begin{tabular}{| l | c c c c c c c c c c |}
\hline
\multicolumn{11}{|c|}{\textbf{arXiv Distribution Distances to Real Text}}\\
\hline
\textbf{LLaMA 13B}& & & & & & & & & & \\
\hline
Sampling & k=30 & k=40 & k=50 & p=0.9 & p=0.95 & p=0.99 & t=0.5 & t=0.7 & t=0.9 & burst\\
\hline
k burst & 0.41 & 0.36 & 0.32 & 0.22 & 0.11 & 0.08 & 0.59 & 0.43 & 0.17 & 0.21 \\
p burst & 0.25 & 0.19 & 0.16 & 0.27 & 0.08 & 0.18 & 0.93 & 0.77 & 0.19 & 0.65 \\
top-p burst & 0.38 & 0.35 & 0.33 & 0.19 & 0.19 & 0.23 & 0.59 & 0.54 & 0.39 & 0.78 \\
log-likelihood & 0.62 & 0.56 & 0.5 & 0.52 & 0.27 & 0.06 & 0.96 & 0.87 & 0.35 & 0.67 \\
log rank & 0.69 & 0.62 & 0.54 & 0.49 & 0.21 & 0.12 & 0.98 & 0.91 & 0.34 & 0.46 \\
rank & 0.52 & 0.51 & 0.5 & 0.42 & 0.32 & 0.07 & 0.55 & 0.5 & 0.19 & 0.33 \\
perplexity & 0.62 & 0.56 & 0.5 & 0.52 & 0.27 & 0.06 & 0.96 & 0.87 & 0.35 & 0.67 \\
diversity & 0.35 & 0.31 & 0.28 & 0.33 & 0.22 & 0.1 & 0.76 & 0.59 & 0.28 & 0.09 \\
recov top k=40 & 0.95 & 0.94 & 0.87 & 0.57 & 0.28 & 0.08 & 0.93 & 0.84 & 0.33 & 0.24 \\
recov top k=50 & 0.93 & 0.93 & 0.92 & 0.58 & 0.3 & 0.08 & 0.91 & 0.83 & 0.33 & 0.25 \\
\hline
\textbf{Vicuna 13B}& & & & & & & & & & \\
\hline
k burst & 0.67 & 0.66 & 0.66 & 0.57 & 0.66 & 0.57 & 0.75 & 0.73 & 0.63 & 0.47 \\
p burst & 0.9 & 0.89 & 0.88 & 0.86 & 0.91 & 0.86 & 0.98 & 0.97 & 0.92 & 0.74 \\
top-p burst & 0.12 & 0.12 & 0.12 & 0.13 & 0.06 & 0.13 & 0.31 & 0.2 & 0.16 & 0.26 \\
log-likelihood & 0.93 & 0.92 & 0.92 & 0.89 & 0.94 & 0.89 & 0.98 & 0.97 & 0.93 & 0.77 \\
log rank & 0.92 & 0.91 & 0.9 & 0.86 & 0.92 & 0.86 & 0.98 & 0.96 & 0.91 & 0.73 \\
rank & 0.45 & 0.44 & 0.44 & 0.38 & 0.44 & 0.38 & 0.49 & 0.47 & 0.4 & 0.18 \\
perplexity & 0.93 & 0.92 & 0.92 & 0.89 & 0.94 & 0.89 & 0.98 & 0.97 & 0.93 & 0.77 \\
diversity & 0.42 & 0.4 & 0.4 & 0.37 & 0.45 & 0.37 & 0.63 & 0.57 & 0.43 & 0.24 \\
recov top k=40 & 0.9 & 0.9 & 0.89 & 0.8 & 0.88 & 0.8 & 0.94 & 0.93 & 0.87 & 0.66 \\
recov top k=50 & 0.89 & 0.89 & 0.88 & 0.79 & 0.87 & 0.79 & 0.93 & 0.92 & 0.85 & 0.63 \\
\hline
\end{tabular}
\caption{\label{arxiv_sep}Metric distribution separability for the arXiv dataset, for each sampling method and each model. Each entry is the result of a KS test between the metrics of the generated text and that same metric over the corresponding real text. 1.0 represents completely separate distributions, while 0.0 represents identical distributions.}
\end{center}
\end{table*}

\begin{table*}
\small
\begin{center}
\begin{tabular}{| l | c c c c c c c c c c |}
\hline
\multicolumn{11}{|c|}{\textbf{CNN Daily Mail Distribution Distances to Real Text}}\\
\hline
\textbf{LLaMA 13B}& & & & & & & & & & \\
\hline
Sampling & k=30 & k=40 & k=50 & p=0.9 & p=0.95 & p=0.99 & t=0.5 & t=0.7 & t=0.9 & burst\\
\hline
k burst & 0.27 & 0.25 & 0.22 & 0.22 & 0.09 & 0.22 & 0.45 & 0.32 & 0.02 & 0.17 \\
p burst & 0.19 & 0.23 & 0.26 & 0.45 & 0.27 & 0.45 & 0.78 & 0.44 & 0.18 & 0.71 \\
top-p burst & 0.56 & 0.53 & 0.51 & 0.44 & 0.42 & 0.44 & 0.85 & 0.79 & 0.6 & 0.85 \\
log-likelihood & 0.12 & 0.04 & 0.04 & 0.43 & 0.17 & 0.43 & 0.87 & 0.61 & 0.12 & 0.82 \\
log rank & 0.21 & 0.1 & 0.04 & 0.44 & 0.19 & 0.44 & 0.93 & 0.69 & 0.11 & 0.58 \\
rank & 0.11 & 0.1 & 0.09 & 0.34 & 0.16 & 0.34 & 0.17 & 0.13 & 0.23 & 0.58 \\
perplexity & 0.12 & 0.04 & 0.04 & 0.43 & 0.17 & 0.43 & 0.87 & 0.61 & 0.12 & 0.82 \\
diversity & 0.46 & 0.5 & 0.53 & 0.69 & 0.58 & 0.69 & 0.26 & 0.16 & 0.54 & 0.62 \\
recov top k=40 & 0.72 & 0.69 & 0.55 & 0.38 & 0.12 & 0.38 & 0.78 & 0.6 & 0.09 & 0.34 \\
recov top k=50 & 0.68 & 0.67 & 0.64 & 0.38 & 0.11 & 0.38 & 0.73 & 0.58 & 0.09 & 0.36 \\
\hline
\textbf{Vicuna 13B}& & & & & & & & & & \\
\hline
k burst & 0.34 & 0.33 & 0.33 & 0.35 & 0.35 & 0.3 & 0.45 & 0.41 & 0.33 & 0.23 \\
p burst & 0.82 & 0.81 & 0.8 & 0.85 & 0.85 & 0.79 & 0.97 & 0.93 & 0.85 & 0.68 \\
top-p burst & 0.34 & 0.34 & 0.34 & 0.32 & 0.32 & 0.33 & 0.61 & 0.48 & 0.39 & 0.41 \\
log-likelihood & 0.86 & 0.85 & 0.85 & 0.88 & 0.88 & 0.84 & 0.96 & 0.94 & 0.87 & 0.73 \\
log rank & 0.81 & 0.79 & 0.79 & 0.83 & 0.83 & 0.76 & 0.95 & 0.91 & 0.82 & 0.63 \\
rank & 0.16 & 0.15 & 0.15 & 0.16 & 0.16 & 0.14 & 0.19 & 0.17 & 0.15 & 0.18 \\
perplexity & 0.86 & 0.85 & 0.85 & 0.88 & 0.88 & 0.84 & 0.96 & 0.94 & 0.87 & 0.73 \\
diversity & 0.35 & 0.37 & 0.38 & 0.31 & 0.31 & 0.38 & 0.11 & 0.19 & 0.32 & 0.47 \\
recov top k=40 & 0.73 & 0.71 & 0.7 & 0.7 & 0.7 & 0.61 & 0.82 & 0.78 & 0.67 & 0.47 \\
recov top k=50 & 0.69 & 0.68 & 0.68 & 0.67 & 0.67 & 0.58 & 0.78 & 0.74 & 0.64 & 0.44 \\
\hline
\end{tabular}
\caption{\label{cnn_dailymail_sep}Metric distribution separability for the CNN Daily Mail dataset, for each sampling method and each model. Each entry is the result of a KS test between the metrics of the generated text and that same metric over the corresponding real text. 1.0 represents completely separate distributions, while 0.0 represents identical distributions.}
\end{center}
\end{table*}

\begin{table*}
\small
\begin{center}
\begin{tabular}{| l | c c c c c c c c c c |}
\hline
\multicolumn{11}{|c|}{\textbf{PG19 Distribution Distances to Real Text}}\\
\hline
\textbf{LLaMA 13B}& & & & & & & & & & \\
\hline
Sampling & k=30 & k=40 & k=50 & p=0.9 & p=0.95 & p=0.99 & t=0.5 & t=0.7 & t=0.9 & burst\\
\hline
k burst & 0.51 & 0.47 & 0.44 & 0.17 & 0.17 & 0.09 & 0.78 & 0.59 & 0.32 & 0.36 \\
p burst & 0.33 & 0.26 & 0.22 & 0.11 & 0.11 & 0.14 & 0.94 & 0.81 & 0.25 & 0.61 \\
top-p burst & 0.45 & 0.43 & 0.4 & 0.26 & 0.26 & 0.31 & 0.49 & 0.57 & 0.45 & 0.88 \\
log-likelihood & 0.71 & 0.65 & 0.6 & 0.36 & 0.36 & 0.07 & 0.97 & 0.9 & 0.43 & 0.58 \\
log rank & 0.76 & 0.69 & 0.64 & 0.27 & 0.27 & 0.08 & 0.98 & 0.92 & 0.39 & 0.4 \\
rank & 0.57 & 0.56 & 0.54 & 0.32 & 0.32 & 0.06 & 0.61 & 0.55 & 0.2 & 0.41 \\
perplexity & 0.71 & 0.65 & 0.6 & 0.36 & 0.36 & 0.07 & 0.97 & 0.9 & 0.43 & 0.58 \\
diversity & 0.62 & 0.6 & 0.57 & 0.5 & 0.5 & 0.41 & 0.86 & 0.74 & 0.54 & 0.43 \\
recov top k=40 & 0.94 & 0.93 & 0.88 & 0.3 & 0.3 & 0.07 & 0.95 & 0.87 & 0.38 & 0.24 \\
recov top k=50 & 0.93 & 0.93 & 0.91 & 0.3 & 0.3 & 0.07 & 0.94 & 0.86 & 0.37 & 0.25 \\
\hline
\textbf{Vicuna 13B}& & & & & & & & & & \\
\hline
k burst & 0.72 & 0.71 & 0.69 & 0.73 & 0.68 & 0.62 & 0.83 & 0.79 & 0.69 & 0.55 \\
p burst & 0.93 & 0.92 & 0.91 & 0.94 & 0.92 & 0.87 & 0.98 & 0.97 & 0.93 & 0.78 \\
top-p burst & 0.14 & 0.13 & 0.13 & 0.06 & 0.08 & 0.13 & 0.29 & 0.2 & 0.16 & 0.29 \\
log-likelihood & 0.96 & 0.95 & 0.95 & 0.96 & 0.95 & 0.92 & 0.98 & 0.98 & 0.95 & 0.82 \\
log rank & 0.95 & 0.95 & 0.94 & 0.95 & 0.93 & 0.88 & 0.98 & 0.97 & 0.94 & 0.78 \\
rank & 0.6 & 0.6 & 0.59 & 0.6 & 0.58 & 0.53 & 0.62 & 0.61 & 0.56 & 0.34 \\
perplexity & 0.96 & 0.95 & 0.95 & 0.96 & 0.95 & 0.92 & 0.98 & 0.98 & 0.95 & 0.82 \\
diversity & 0.63 & 0.62 & 0.62 & 0.66 & 0.63 & 0.57 & 0.76 & 0.72 & 0.64 & 0.47 \\
recov top k=40 & 0.95 & 0.94 & 0.94 & 0.93 & 0.9 & 0.83 & 0.96 & 0.95 & 0.9 & 0.7 \\
recov top k=50 & 0.94 & 0.94 & 0.94 & 0.93 & 0.89 & 0.82 & 0.95 & 0.95 & 0.89 & 0.69 \\
\hline
\end{tabular}
\caption{\label{pg19_sep}Metric distribution separability for the PG19 dataset, for each sampling method and each model. Each entry is the result of a KS test between the metrics of the generated text and that same metric over the corresponding real text. 1.0 represents completely separate distributions, while 0.0 represents identical distributions.}
\end{center}
\end{table*}

\begin{table*}
\small
\begin{center}
\begin{tabular}{| l | c c c c c c c c c c |}
\hline
\multicolumn{11}{|c|}{\textbf{StackExchange Distribution Distances to Real Text}}\\
\hline
\textbf{LLaMA 13B}& & & & & & & & & & \\
\hline
Sampling & k=30 & k=40 & k=50 & p=0.9 & p=0.95 & p=0.99 & t=0.5 & t=0.7 & t=0.9 & burst\\
\hline
k burst & 0.16 & 0.13 & 0.11 & 0.08 & 0.11 & 0.26 & 0.31 & 0.21 & 0.12 & 0.07 \\
p burst & 0.08 & 0.11 & 0.12 & 0.1 & 0.08 & 0.19 & 0.57 & 0.36 & 0.07 & 0.52 \\
top-p burst & 0.27 & 0.26 & 0.26 & 0.18 & 0.2 & 0.25 & 0.42 & 0.37 & 0.28 & 0.67 \\
log-likelihood & 0.24 & 0.18 & 0.15 & 0.21 & 0.09 & 0.19 & 0.65 & 0.47 & 0.13 & 0.62 \\
log rank & 0.26 & 0.19 & 0.15 & 0.19 & 0.07 & 0.23 & 0.76 & 0.53 & 0.11 & 0.52 \\
rank & 0.23 & 0.12 & 0.11 & 0.08 & 0.06 & 0.23 & 0.18 & 0.13 & 0.11 & 0.55 \\
perplexity & 0.24 & 0.18 & 0.15 & 0.21 & 0.09 & 0.19 & 0.65 & 0.47 & 0.13 & 0.62 \\
diversity & 0.14 & 0.16 & 0.18 & 0.13 & 0.19 & 0.29 & 0.33 & 0.13 & 0.16 & 0.27 \\
recov top k=40 & 0.67 & 0.58 & 0.45 & 0.28 & 0.06 & 0.28 & 0.61 & 0.5 & 0.07 & 0.42 \\
recov top k=50 & 0.63 & 0.56 & 0.53 & 0.28 & 0.06 & 0.29 & 0.57 & 0.46 & 0.08 & 0.45 \\
\hline
\textbf{Vicuna 13B}& & & & & & & & & & \\
\hline
k burst & 0.21 & 0.2 & 0.2 & 0.27 & 0.23 & 0.19 & 0.32 & 0.29 & 0.22 & 0.09 \\
p burst & 0.6 & 0.58 & 0.59 & 0.72 & 0.67 & 0.59 & 0.86 & 0.8 & 0.66 & 0.44 \\
top-p burst & 0.15 & 0.16 & 0.16 & 0.12 & 0.12 & 0.14 & 0.24 & 0.18 & 0.16 & 0.24 \\
log-likelihood & 0.7 & 0.68 & 0.69 & 0.8 & 0.76 & 0.68 & 0.88 & 0.84 & 0.73 & 0.51 \\
log rank & 0.66 & 0.64 & 0.65 & 0.75 & 0.71 & 0.63 & 0.86 & 0.81 & 0.69 & 0.46 \\
rank & 0.32 & 0.31 & 0.31 & 0.34 & 0.33 & 0.3 & 0.36 & 0.35 & 0.31 & 0.17 \\
perplexity & 0.7 & 0.68 & 0.69 & 0.8 & 0.76 & 0.68 & 0.88 & 0.84 & 0.73 & 0.51 \\
diversity & 0.22 & 0.21 & 0.22 & 0.31 & 0.27 & 0.22 & 0.39 & 0.34 & 0.26 & 0.13 \\
recov top k=40 & 0.69 & 0.68 & 0.67 & 0.71 & 0.69 & 0.62 & 0.75 & 0.73 & 0.66 & 0.45 \\
recov top k=50 & 0.63 & 0.62 & 0.63 & 0.66 & 0.64 & 0.57 & 0.7 & 0.68 & 0.6 & 0.4 \\
\hline
\end{tabular}
\caption{\label{stackexchange_sep}Metric distribution separability for the StackExchange dataset, for each sampling method and each model. Each entry is the result of a KS test between the metrics of the generated text and that same metric over the corresponding real text. 1.0 represents completely separate distributions, while 0.0 represents identical distributions.}
\end{center}
\end{table*}

\begin{table*}
\small
\begin{center}
\begin{tabular}{| l | c c c c c c c c c c |}
\hline
\multicolumn{11}{|c|}{\textbf{Twitter Distribution Distances to Real Text}}\\
\hline
\textbf{LLaMA 13B}& & & & & & & & & & \\
\hline
Sampling & k=30 & k=40 & k=50 & p=0.9 & p=0.95 & p=0.99 & t=0.5 & t=0.7 & t=0.9 & burst\\
\hline
k burst & 0.71 & 0.68 & 0.66 & 0.62 & 0.57 & 0.56 & 0.94 & 0.83 & 0.71 & 0.53 \\
p burst & 0.67 & 0.63 & 0.6 & 0.61 & 0.51 & 0.35 & 0.96 & 0.89 & 0.6 & 0.68 \\
top-p burst & 0.13 & 0.12 & 0.12 & 0.21 & 0.2 & 0.17 & 0.12 & 0.19 & 0.15 & 0.77 \\
log-likelihood & 0.86 & 0.84 & 0.84 & 0.8 & 0.72 & 0.57 & 0.98 & 0.93 & 0.77 & 0.61 \\
log rank & 0.85 & 0.83 & 0.81 & 0.73 & 0.63 & 0.46 & 0.97 & 0.92 & 0.7 & 0.56 \\
rank & 0.57 & 0.56 & 0.56 & 0.5 & 0.43 & 0.25 & 0.63 & 0.58 & 0.37 & 0.64 \\
perplexity & 0.86 & 0.84 & 0.84 & 0.8 & 0.72 & 0.57 & 0.98 & 0.93 & 0.77 & 0.61 \\
diversity & 0.85 & 0.85 & 0.84 & 0.82 & 0.8 & 0.78 & 0.95 & 0.91 & 0.83 & 0.7 \\
recov top k=40 & 0.89 & 0.88 & 0.86 & 0.71 & 0.6 & 0.41 & 0.94 & 0.89 & 0.66 & 0.46 \\
recov top k=50 & 0.88 & 0.87 & 0.86 & 0.7 & 0.59 & 0.39 & 0.94 & 0.87 & 0.64 & 0.49 \\
\hline
\textbf{Vicuna 13B}& & & & & & & & & & \\
\hline
k burst & 0.83 & 0.82 & 0.82 & 0.8 & 0.83 & 0.8 & 0.9 & 0.88 & 0.84 & 0.72 \\
p burst & 0.9 & 0.89 & 0.89 & 0.87 & 0.89 & 0.87 & 0.95 & 0.94 & 0.9 & 0.73 \\
top-p burst & 0.13 & 0.14 & 0.13 & 0.12 & 0.17 & 0.12 & 0.07 & 0.11 & 0.12 & 0.23 \\
log-likelihood & 0.93 & 0.93 & 0.92 & 0.9 & 0.92 & 0.9 & 0.96 & 0.95 & 0.92 & 0.81 \\
log rank & 0.91 & 0.91 & 0.9 & 0.87 & 0.89 & 0.87 & 0.95 & 0.94 & 0.9 & 0.76 \\
rank & 0.59 & 0.6 & 0.6 & 0.56 & 0.59 & 0.56 & 0.62 & 0.61 & 0.57 & 0.32 \\
perplexity & 0.93 & 0.93 & 0.92 & 0.9 & 0.92 & 0.9 & 0.96 & 0.95 & 0.92 & 0.81 \\
diversity & 0.8 & 0.8 & 0.8 & 0.78 & 0.79 & 0.78 & 0.86 & 0.83 & 0.8 & 0.68 \\
recov top k=40 & 0.9 & 0.9 & 0.89 & 0.84 & 0.86 & 0.84 & 0.92 & 0.91 & 0.87 & 0.71 \\
recov top k=50 & 0.89 & 0.89 & 0.89 & 0.83 & 0.86 & 0.83 & 0.91 & 0.9 & 0.86 & 0.69 \\
\hline
\end{tabular}
\caption{\label{twitter_sep}Metric distribution separability for the Twitter dataset, for each sampling method and each model. Each entry is the result of a KS test between the metrics of the generated text and that same metric over the corresponding real text. 1.0 represents completely separate distributions, while 0.0 represents identical distributions.}
\end{center}
\end{table*}

\begin{table*}
\small
\begin{center}
\begin{tabular}{| l | c c c c c c c c c c |}
\hline
\multicolumn{11}{|c|}{\textbf{Wikipedia Distribution Distances to Real Text}}\\
\hline
\textbf{LLaMA 13B}& & & & & & & & & & \\
\hline
Sampling & k=30 & k=40 & k=50 & p=0.9 & p=0.95 & p=0.99 & t=0.5 & t=0.7 & t=0.9 & burst\\
\hline
k burst & 0.37 & 0.35 & 0.34 & 0.24 & 0.24 & 0.12 & 0.46 & 0.38 & 0.23 & 0.3 \\
p burst & 0.06 & 0.09 & 0.12 & 0.1 & 0.1 & 0.23 & 0.56 & 0.34 & 0.04 & 0.54 \\
top-p burst & 0.35 & 0.33 & 0.33 & 0.24 & 0.24 & 0.28 & 0.46 & 0.41 & 0.34 & 0.71 \\
log-likelihood & 0.17 & 0.12 & 0.09 & 0.07 & 0.07 & 0.17 & 0.66 & 0.49 & 0.09 & 0.64 \\
log rank & 0.21 & 0.16 & 0.11 & 0.03 & 0.03 & 0.21 & 0.73 & 0.54 & 0.07 & 0.47 \\
rank & 0.09 & 0.09 & 0.09 & 0.16 & 0.16 & 0.29 & 0.08 & 0.09 & 0.22 & 0.56 \\
perplexity & 0.17 & 0.12 & 0.09 & 0.07 & 0.07 & 0.17 & 0.66 & 0.49 & 0.09 & 0.64 \\
diversity & 0.08 & 0.1 & 0.12 & 0.14 & 0.14 & 0.22 & 0.31 & 0.14 & 0.11 & 0.19 \\
recov top k=40 & 0.63 & 0.61 & 0.52 & 0.13 & 0.13 & 0.15 & 0.61 & 0.51 & 0.14 & 0.24 \\
recov top k=50 & 0.6 & 0.59 & 0.58 & 0.13 & 0.13 & 0.15 & 0.58 & 0.49 & 0.12 & 0.27 \\
\hline
\textbf{Vicuna 13B}& & & & & & & & & & \\
\hline
k burst & 0.43 & 0.43 & 0.43 & 0.4 & 0.43 & 0.4 & 0.49 & 0.47 & 0.42 & 0.36 \\
p burst & 0.68 & 0.68 & 0.68 & 0.66 & 0.72 & 0.66 & 0.86 & 0.81 & 0.72 & 0.58 \\
top-p burst & 0.2 & 0.2 & 0.19 & 0.17 & 0.17 & 0.17 & 0.39 & 0.3 & 0.21 & 0.25 \\
log-likelihood & 0.76 & 0.76 & 0.75 & 0.74 & 0.78 & 0.74 & 0.87 & 0.83 & 0.77 & 0.65 \\
log rank & 0.71 & 0.7 & 0.69 & 0.67 & 0.71 & 0.67 & 0.84 & 0.79 & 0.71 & 0.58 \\
rank & 0.09 & 0.08 & 0.08 & 0.08 & 0.08 & 0.08 & 0.11 & 0.09 & 0.08 & 0.13 \\
perplexity & 0.76 & 0.76 & 0.75 & 0.74 & 0.78 & 0.74 & 0.87 & 0.83 & 0.77 & 0.65 \\
diversity & 0.06 & 0.06 & 0.06 & 0.05 & 0.07 & 0.05 & 0.18 & 0.13 & 0.07 & 0.04 \\
recov top k=40 & 0.66 & 0.66 & 0.64 & 0.57 & 0.62 & 0.57 & 0.7 & 0.67 & 0.61 & 0.49 \\
recov top k=50 & 0.64 & 0.63 & 0.62 & 0.55 & 0.6 & 0.55 & 0.67 & 0.64 & 0.59 & 0.46 \\
\hline
\end{tabular}
\caption{\label{wikipedia_sep}Metric distribution separability for the Wikipedia dataset, for each sampling method and each model. Each entry is the result of a KS test between the metrics of the generated text and that same metric over the corresponding real text. 1.0 represents completely separate distributions, while 0.0 represents identical distributions.}
\end{center}
\end{table*}

\begin{table*}
\small
\begin{center}
\begin{tabular}{| l | c c c c c c c c c c |}
\hline
\multicolumn{11}{|c|}{\textbf{GLTR Logistic Regression F1 Scores}}\\
\hline
\textbf{LLaMA 13B}& & & & & & & & & & \\
\hline
Sampling: & k=30 & k=40 & k=50 & p=0.9 & p=0.95 & p=0.99 & t=0.5 & t=0.7 & t=0.9 & burst\\
\hline
ArXiv & 0.88 & 0.89 & 0.9 & 0.77 & 0.64 & 0.59 & 0.98 & 0.94 & 0.66 & 0.6\\
CNN DailyMail & 0.72 & 0.77 & 0.79 & 0.61 & 0.6 & 0.71 & 0.96 & 0.86 & 0.57 & 0.71\\
PG-19 & 0.93 & 0.92 & 0.91 & 0.78 & 0.65 & 0.56 & 0.99 & 0.95 & 0.69 & 0.67\\
StackExchange & 0.69 & 0.65 & 0.67 & 0.58 & 0.5 & 0.65 & 0.9 & 0.79 & 0.57 & 0.75\\
Twitter & 0.92 & 0.91 & 0.91 & 0.86 & 0.82 & 0.72 & 0.98 & 0.95 & 0.84 & 0.78\\
Wikipedia & 0.7 & 0.71 & 0.72 & 0.63 & 0.53 & 0.6 & 0.87 & 0.78 & 0.58 & 0.68\\
\hline
\textbf{Vicuna 13B}& & & & & & & & & & \\
\hline
ArXiv & 0.96 & 0.96 & 0.95 & 0.97 & 0.96 & 0.93 & 0.99 & 0.97 & 0.96 & 0.88\\
CNN DailyMail & 0.92 & 0.9 & 0.9 & 0.95 & 0.93 & 0.9 & 0.97 & 0.96 & 0.92 & 0.86\\
PG-19 & 0.97 & 0.97 & 0.97 & 0.97 & 0.96 & 0.94 & 0.99 & 0.99 & 0.97 & 0.89\\
StackExchange & 0.82 & 0.81 & 0.81 & 0.87 & 0.84 & 0.82 & 0.9 & 0.89 & 0.83 & 0.72\\
Twitter & 0.95 & 0.95 & 0.95 & 0.95 & 0.94 & 0.93 & 0.97 & 0.96 & 0.94 & 0.88\\
Wikipedia & 0.86 & 0.85 & 0.84 & 0.88 & 0.85 & 0.85 & 0.91 & 0.89 & 0.86 & 0.8\\
\hline
\end{tabular}
\caption{\label{gltr_f1}F1 scores for logistic regression classifiers trained to classify real vs synthetic text, using GLTR bins as features. A high score indicates that based on the GLTR bins, it was easier to detect if a piece of text was real or generated.}
\end{center}
\end{table*}

\begin{table*}
\small
\begin{center}
\begin{tabular}{| l | c c c c c c c c c c |}
\hline
\multicolumn{11}{|c|}{\textbf{All-Metric Logistic Regression F1 Scores}}\\
\hline
\textbf{LLaMa 13B}& & & & & & & & & & \\
\hline
Sampling: & k=30 & k=40 & k=50 & p=0.9 & p=0.95 & p=0.99 & t=0.5 & t=0.7 & t=0.9 & burst\\
\hline
ArXiv & 0.97 & 0.95 & 0.96 & 0.94 & 0.88 & 0.75 & 0.99 & 0.99 & 0.86 & 0.97\\
CNN DailyMail & 0.93 & 0.93 & 0.92 & 0.95 & 0.91 & 0.87 & 0.99 & 0.98 & 0.92 & 0.99\\
PG-19 & 0.98 & 0.97 & 0.96 & 0.95 & 0.91 & 0.83 & 0.99 & 0.99 & 0.91 & 0.98\\
StackExchange & 0.91 & 0.87 & 0.87 & 0.91 & 0.87 & 0.8 & 0.98 & 0.97 & 0.87 & 0.97\\
Twitter & 0.97 & 0.97 & 0.96 & 0.95 & 0.93 & 0.91 & 0.99 & 0.98 & 0.95 & 0.94\\
Wikipedia & 0.84 & 0.83 & 0.83 & 0.86 & 0.81 & 0.73 & 0.96 & 0.91 & 0.83 & 0.94\\
\hline
\textbf{Vicuna 13B}& & & & & & & & & & \\
\hline
ArXiv & 0.99 & 0.99 & 0.99 & 1.0 & 0.99 & 0.99 & 1.0 & 0.99 & 0.99 & 0.98\\
CNN DailyMail & 0.99 & 0.99 & 0.99 & 0.99 & 0.99 & 0.99 & 1.0 & 1.0 & 0.99 & 0.99\\
PG-19 & 0.99 & 0.99 & 0.99 & 1.0 & 1.0 & 0.99 & 1.0 & 1.0 & 1.0 & 0.98\\
StackExchange & 0.99 & 0.98 & 0.98 & 0.99 & 0.99 & 0.99 & 0.99 & 0.99 & 0.99 & 0.97\\
Twitter & 0.98 & 0.98 & 0.98 & 0.98 & 0.98 & 0.97 & 0.99 & 0.98 & 0.98 & 0.94\\
Wikipedia & 0.97 & 0.97 & 0.97 & 0.98 & 0.97 & 0.97 & 0.99 & 0.98 & 0.98 & 0.97\\
\hline
\end{tabular}
\caption{\label{logreg_f1}F1 scores for logistic regression classifiers trained to classify real vs synthetic text, using all metrics as input features. A high score indicates that it was easier to detect if a piece of text was real or generated.}
\end{center}
\end{table*}

\end{document}